%% file: main.tex
\documentclass{article}

\usepackage{arxiv}

\usepackage[utf8]{inputenc} 
\usepackage[T1]{fontenc}    
\usepackage{hyperref}       
\usepackage{url}            
\usepackage{booktabs}       
\usepackage{amsfonts}       
\usepackage{nicefrac}       
\usepackage{microtype}      
\usepackage[ruled]{algorithm2e}
\usepackage{rotating}
\usepackage{graphicx}
\usepackage{multirow}
\usepackage{algorithmic}
\usepackage{amsmath}
\usepackage[]{color}
\usepackage{subfig}

\title{How Complex is your classification problem?\\ A survey on measuring classification complexity} 

\author{Ana C. Lorena\\
Instituto Tecnol\'ogico de Aeron\'autica\\
S\~ao Jos\'e dos Campos, SP, Brazil\\
\texttt{aclorena@ita.br}\\
\And
Lu\'is P. F. Garcia\\
Universidade de Bras\'ilia\\
Bras\'ilia, DF, Brazil\\
\texttt{luis.garcia@unb.br}\\
\And
Jens Lehmann\\
University of Bonn\\
Bonn, Germany\\
\texttt{jens.lehmann@cs.uni-bonn.de}\\
\And
Marcilio C. P. Souto\\
University of Orl\'eans\\
Orleans, France\\
\texttt{marcilio.desouto@univ-orleans.fr}\\
\And
Tin K. Ho\\
IBM Watson\\
Yorktown Heights, USA\\
\texttt{tho@us.ibm.com}\\
}

\begin{document}
\maketitle

\begin{abstract}
Characteristics extracted from the training datasets of classification problems have proven to be effective predictors in a number of meta-analyses. Among them, measures of classification complexity can be used to estimate the difficulty in separating the data points into their expected classes. Descriptors of the spatial distribution of the data and estimates of the shape and size of the decision boundary are among the known measures for this characterization. This information can support the formulation of new data-driven pre-processing and pattern recognition techniques, which can in turn be focused on challenges highlighted by such characteristics of the problems.  This paper surveys and analyzes measures which can be extracted from the 
training datasets 
in order to characterize the complexity of the respective classification problems. Their use in recent literature is also reviewed and discussed, 
allowing to prospect opportunities for future work in the area. Finally, descriptions are given on an R package named Extended Complexity Library (ECoL) that implements a set of complexity measures and is made publicly available.
\end{abstract}


\maketitle

\input{Introduction}
\input{ComplexityMeasures}

\input{Others}

\input{ECol}
\input{Aplications}

\input{Conclusion}

\bibliographystyle{plainnat}
\bibliography{ref}

\end{document}

%% file: Introduction.tex
\section{Introduction} \label{cap:introducao}

The work from \citet{HoBasu2002} was seminal in analyzing the difficulty of a classification problem by using descriptors extracted from a learning dataset. Given that no Machine Learning (ML) technique can consistently obtain the best performance for every classification problem \citep{wolpert1996lack}, this type of analysis allows to understand the scenarios in which a given ML technique succeeds and fails  \citep{ali2006learning,flores2014domains,luengo2015automatic,munoz2018instance}. Furthermore, it guides the development of new data-driven pre-processing and pattern recognition techniques, as done in \citep{DongKothari2003,smith2014instance,MollinedaEtAl2005,hu2010selecting,GarciaEtAl2015}. This data-driven approach enables a better understanding of the peculiarities of a given application domain that can be explored in order to get better prediction results.  

According to \citet{HoBasu2002}, the complexity of a classification problem can be attributed to a combination of three main factors: (i) the ambiguity of the classes; (ii) the sparsity and dimensionality of the data; and (iii) the complexity of the boundary separating the classes. The ambiguity of the classes is present in scenarios in which the classes can not be distinguished using the data provided, regardless of the classification algorithm employed. This is the case for poorly defined concepts and the use of non-discriminative data features. These problems are known to have non-zero Bayes error. An incomplete or sparse dataset also hinders a proper data analysis. This shortage leads to some input space regions to be underconstrained.  After training, subsequent data residing in those regions are classified arbitrarily. 
Finally, \citet{HoBasu2002} focus on the complexity of the classification boundary, and present a number of measures that characterize the boundary in different ways. The complexity of classification boundary is related to the size of the smallest description needed to represent the classes and is native of the problem itself \citep{Antolinez2011}.  Using the Kolmogorov complexity concept \citep{Vitanyi1993}, the complexity of a classification problem can be measured by the size of the smallest algorithm which is able to describe the relationships between the data  \citep{LiAbu-Mostafa2006}. In the worst case, it would be necessary to list all the objects along with their labels. However, if there is some regularity in the data, a compact algorithm
can be obtained. In practice, the Kolmogorov complexity is incomputable and approximations are made, as those based on the computation of 
indicators and geometric descriptors 
drawn from the learning datasets available for training a classifer
\citep{HoBasu2002,Singh2003}. 
We refer to those indicators and geometric descriptors as data complexity measures or simply {\em complexity measures} from here on. 

This paper surveys the main complexity measures that can be obtained directly from the data available for learning. It extends the work from \citet{HoBasu2002} by including more measures from literature that may complement the concepts already covered by the measures proposed in their work. The formulations of some of the measures are also adapted so that they can give standardized results. The usage of the complexity measures through recent literature is reviewed too, highlighting various domains where an advantageous use of the measures can be achieved. Besides, the main strengths and weakness of each measure are reported. As a side result, this analysis provides insights into adaptations needed with some of the measures, and into new unexplored areas where the complexity measures can succeed. 

All measures detailed in this survey were assembled into an R package named ECoL (\textit{Extended Complexity Library}). It contains all the measures from the DCoL (\textit{Data Complexity}) library \citep{Orriols-PuigEtAl2010}, which were standardized and reimplemented in R, and a set of novel measures from the related literature. The added measures were chosen in order to complement the concepts assessed by the original complexity measures. Some corrections into the existent measures are also discussed and detailed in the paper. The ECoL package is publicly available at CRAN\footnote{\url{https://cran.r-project.org/package=ECoL}} and GitHub\footnote{\url{https://github.com/lpfgarcia/ECoL}}. 

This paper is structured as follows: Section \ref{cap:medidas} presents the most relevant complexity measures. Section \ref{cap:core} presents and analyzes the complexity measures included in the ECol package. Section \ref{cap:apli} presents some applications of the complexity measures in the ML literature.  
Section \ref{cap:concl} concludes this work. 

%% file: ComplexityMeasures.tex
\section{Complexity Measures} \label{cap:medidas}

Geometric and statistical data descriptors are among the most used in the characterization of the complexity of classification problems. Among them are the measures proposed in \citep{HoBasu2002} to describe the complexity of the boundary needed to separate binary classification problems, later extended to multiclass classification problems in works like \citep{MollinedaEtAl2005,HoBasuLaw2006,sotoca2006meta,Orriols-PuigEtAl2010}. \citet{HoBasu2002} divide their measures into three main groups: (i) measures of overlap of individual feature values;
(ii) measures of the separability of classes; 
and (iii) geometry, topology and density of manifolds measures. Similarly, \citet{SotocaEtAl2005} divide the complexity measures into the categories: (i) measures of overlap; (ii) measures of class separability; and (iii) measures of geometry and density.  In this paper, we group the complexity measures into more categories, as follows:  

\begin{enumerate}
\item \textbf{Feature-based measures}, which characterize how informative the available features are to separate the classes; 
\item \textbf{Linearity measures}, which try to quantify whether the classes can be linearly separated;
\item \textbf{Neighborhood measures}, which characterize the presence and density of same or different classes in local neighborhoods;
\item \textbf{Network measures}, which extract structural information from the dataset by modeling it as a graph;
\item \textbf{Dimensionality measures}, which evaluate data sparsity based on the number of samples relative to the data dimensionality;
\item \textbf{Class imbalance measures}, which consider the ratio of the numbers of examples between classes.
\end{enumerate}

To define the measures, we consider that they are estimated from a learning dataset $T$ (or part of it) containing $n$ pairs of examples $(\mathbf x_i, y_i)$, where $\mathbf x_i = (x_ {i1}, \ldots, x_ {im})$ and $y_i \in \{1, \ldots, n_c\} $. That is, each example $\mathbf x_i$ is described by $m$ predictive features and has a label $y_i$ out of $n_c$ classes. Most of the measures are defined for features with numerical values only. In this case, symbolic values must be properly converted into numerical values prior to their use. We also use an assumption that 
linearly separable problems can be considered simpler than classification problems requiring non-linear decision boundaries. Finally, some measures are defined for binary classification problems only. In that case, a multiclass problem must be first decomposed into multiple binary sub-problems. Here we adopt a pairwise analysis of the classes, that is, a one-versus-one (OVO) decomposition of the multiclass problem \citep{LorenaEtAl2008}. 
The measure for the multiclass problem is then defined as the average of the values across the different sub-problems. In order to standardize the interpretation of the measure values, we introduce some modifications into the original definitions of some of the measures. The objective was to make each measure assume values in bounded intervals and also to  make higher values of the measures indicative of a higher complexity, whilst lower values indicate a lower complexity.

\subsection{Feature-based Measures}

These measures evaluate the discriminative power of the features.  
In many of them each feature is evaluated individually. If there is at least one very discriminative feature in the dataset, the problem can be considered simpler than if there is no such an attribute. All measures from this category require the features to have numerical values. 
Most of the measures are also defined for binary classification problems only.

\subsubsection{Maximum Fisher's Discriminant Ratio (F1)} 

The first measure presented in this category is the maximum Fisher's discriminant ratio, denoted by F1. It measures the overlap between the values of the features in different classes and is given by:
\begin{equation} \label{eq:F1}
F1 = \frac{1}{1 + \max_{i=1}^{m} r_{f_i}},
\end{equation} 
where $r_{f_i}$ is a discriminant ratio for each feature $f_i$. Originally, F1 takes the value of the largest discriminant ratio among all the available features. This is consistent with the definition that if at least one feature discriminates the classes, the dataset can be considered simpler than if no such attribute exists. In this paper we take the inverse of the original F1 formulation into account, as presented in Equation \ref{eq:F1}. Herewith, the F1 values become bounded in the $(0,1]$ interval and higher values indicate more complex problems, where no individual feature is able to discriminate the classes.

\begin{figure}[ht!]
   \centering
\includegraphics[width=0.60\textwidth]
        {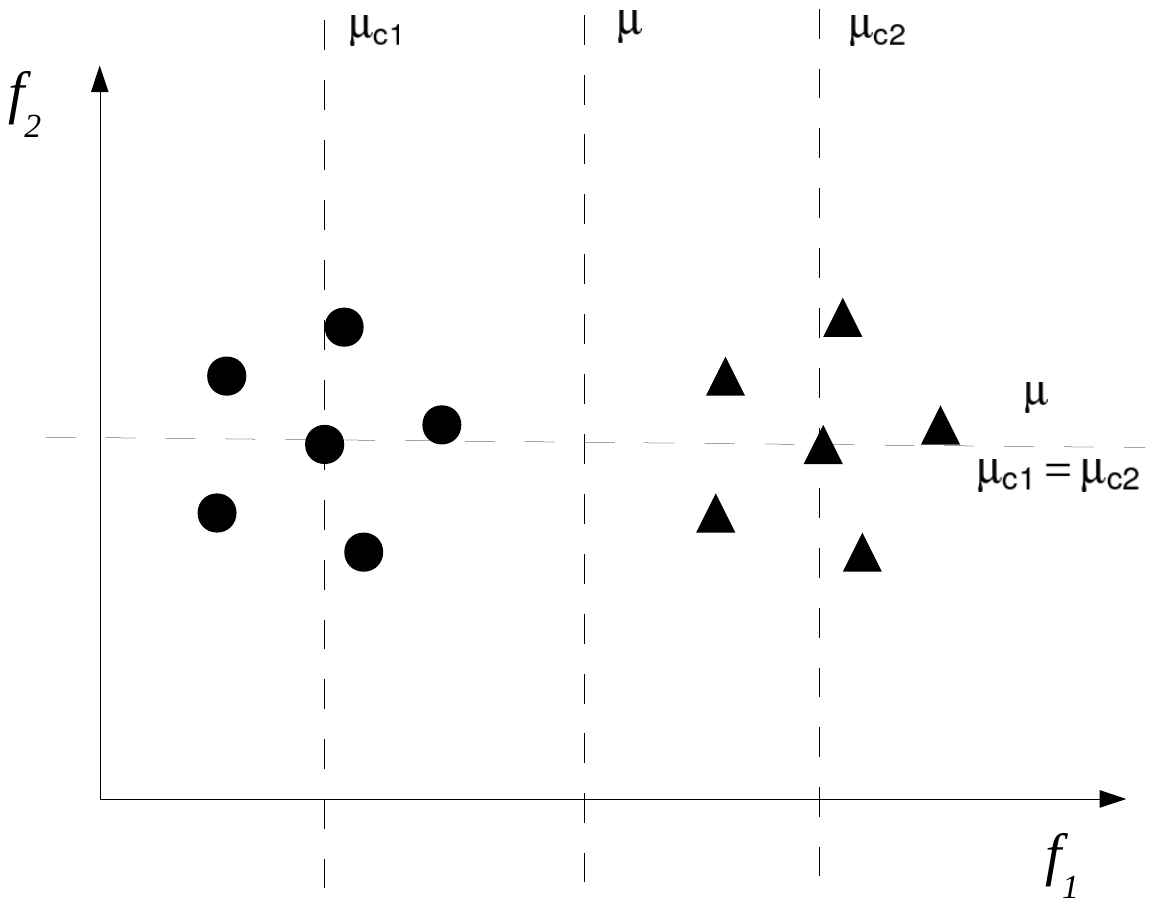}
   \caption{Example of F1 computation for a two-class dataset}
   \label{fig:exF1}       
\end{figure}

\citet{Orriols-PuigEtAl2010} present different equations for calculating $r_{f_i}$, depending on the number of classes or whether the features are continuous or ordinal \citep{Cummins2013}. One straightforward formulation is:
\begin{equation} \label{eq:F1_1}
r_{f_i} = \frac{\sum_{j=1}^{n_c} \sum_{k=1, k \neq j}^{n_c} p_{c_j} p_{c_k} (\mu_{c_j}^{f_i}-\mu_{c_k}^{f_i})^2}{\sum_{j=1}^{n_c} p_{c_j} (\sigma_{c_j}^{f_i})^2},
\end{equation}
where $p_{c_j}$ is the proportion of examples in class $c_j$, $\mu_{c_j}^{f_i}$ is the mean of feature $f_i$ across examples of class $c_j$ and $\sigma_{c_j}^{f_i}$ is the standard deviation of such values. An alternative for $r_{f_i}$ computation which can be employed for both 
binary and multiclass classification problems is given in \citep{MollinedaEtAl2005}. Here we adopt this formulation, which is similar to the clustering validation index from \citet{calinski1974dendrite}: 
\begin{equation} \label{F1multi2}
r_{f_i} = \frac{\sum_{j=1}^{n_c} n_{c_j} (\mu_{c_j}^{f_i}-\mu^{f_i})^2}{\sum_{j=1}^{n_c} \sum_{l=1}^{n_{c_j}} (x_{li}^j-\mu_{c_j}^{f_i})^2},
\end{equation}
where $n_{c_j}$ is the number of examples in class $c_j$, $\mu_{c_j}^{f_i}$ is the same as defined for Equation \ref{eq:F1_1}, 
$\mu^{f_i}$ is the mean 
of the $f_i$ values across all the classes, 
and $x_{li}^j$ denotes the individual value of the feature $f_i$ for an example from class $c_j$. Taking, for instance, the dataset shown in Figure \ref{fig:exF1}, the most discriminative feature would be $f_1$. F1 correctly indicates that the classes can be easily separable using this feature. Feature $f_2$, on the other hand, is non-discriminative, since its values for the two classes overlap, with the same mean and variance. 

The denominator in Equation \ref{F1multi2} must go through all examples in the dataset. The numerator goes through the classes. Since the discriminant ratio must be computed for all features, the total asymptotic cost for the F1 computation is $O(m \cdot (n + n_c))$. As $n \geq n_c$ (there is at least one example per class), $O(m \cdot (n + n_c))$ can be reduced to $O(m \cdot n)$. 

Roughly, the F1 measure computes the ratio of inter-class  
to the intra-class scatter for each feature. Using the formulation in Equation \ref{eq:F1}, low values of the F1 measure indicate that there is at least one feature whose values show little overlap among the different classes; that is, it indicates the existence of a feature for which a hyperplane perpendicular to its axis can separate the classes fairly. Nonetheless, 
if the required hyperplane is oblique to the feature axes, F1 may not be able to reflect the underlying simplicity. In order to deal with this issue, \citet{Orriols-PuigEtAl2010} propose to use a F1 variant based on a {\em Directional Vector}, to be discussed next. 
Finally, \citet{hu2010selecting} note that the F1 measure is most effective if the probability distributions of the classes are approximately normal, which is not always true. On the contrary, there can be highly separable classes, such as those distributed on the surfaces of two concentric hyperspheres, that would yield a very high value for F1. 

\subsubsection{The Directional-vector Maximum Fisher's Discriminant Ratio (F1v)}

This measure is used in \citet{Orriols-PuigEtAl2010} as a complement to F1. It searches for a vector which can separate the two classes after the examples have been projected into it and considers a directional Fisher criterion defined in  \citet{malina2001two} as:
\begin{equation} \label{eq:dF}
dF = \frac{\mathbf d^t \mathbf B \mathbf d}{\mathbf d^t \mathbf W \mathbf d},
\end{equation} 
where $\mathbf{d}$ is the directional vector onto which data are projected in order to maximize class separation, $\mathbf{B}$ is the between-class scatter matrix and $\mathbf{W}$ is the within-class scatter matrix. $\mathbf{d}$, $\mathbf{B}$ and $\mathbf{W}$ are defined according to Equations \ref{eq:d}, \ref{eq:B} and \ref{eq:W}, respectively.
\begin{equation} \label{eq:d}
\mathbf d = \mathbf W^{-1}(\mu_{c_1}-\mu_{c_2}),
\end{equation}
where $\mu_{c_i}$ is the centroid (mean vector) of class $c_i$ and $\mathbf W^{-1}$ is the pseudo-inverse of $\mathbf W$.  
\begin{equation} \label{eq:B}
\mathbf B = (\mu_{c_1}-\mu_{c_2})(\mu_{c_1}-\mu_{c_2})^t
\end{equation} 
\begin{equation} \label{eq:W}
\mathbf W = p_{c_1}\Sigma_{c_1} + p_{c_2}\Sigma_{c_2},
\end{equation} 
where $p_{c_i}$ is the proportion of examples in class $c_i$ and $\Sigma_{c_1}$ is the scatter matrix of  class $c_i$. 

Taking the definition of dF, the F1v measure is given by:
\begin{equation}\label{eq:F1v}
F1v = \frac{1}{1+dF}
\end{equation}

According to \citet{Orriols-PuigEtAl2010}, the asymptotic cost of the F1v algorithm for a binary classification problem is $O(m \cdot n + m^3)$. Multiclass problems are first decomposed according to the OVO strategy, producing $\frac{n_c(n_c-1)}{2}$ subproblems. In the case that each one of them has the same number of examples, that is, $\frac{n}{n_c}$, the total cost of the F1v measure computation is $O(m \cdot n \cdot n_c + m^3 \cdot n_c^2)$.

Lower values in F1v defined by Equation \ref{eq:F1v}, which are bounded in the $(0,1]$ interval, indicate simpler classification problems. In this case, a linear hyperplane will be able to separate most if not all of the data, in a suitable orientation with regard to the features axes.  
This measure can be quite costly to compute due to the need for the pseudo-inverse of the scatter matrix.  Like F1, it is based on the assumption of normality of the classes distributions. 

\subsubsection{Volume of Overlapping Region (F2)}

\begin{figure}[ht!]
	\centering%
	\includegraphics[width=0.5\textwidth]{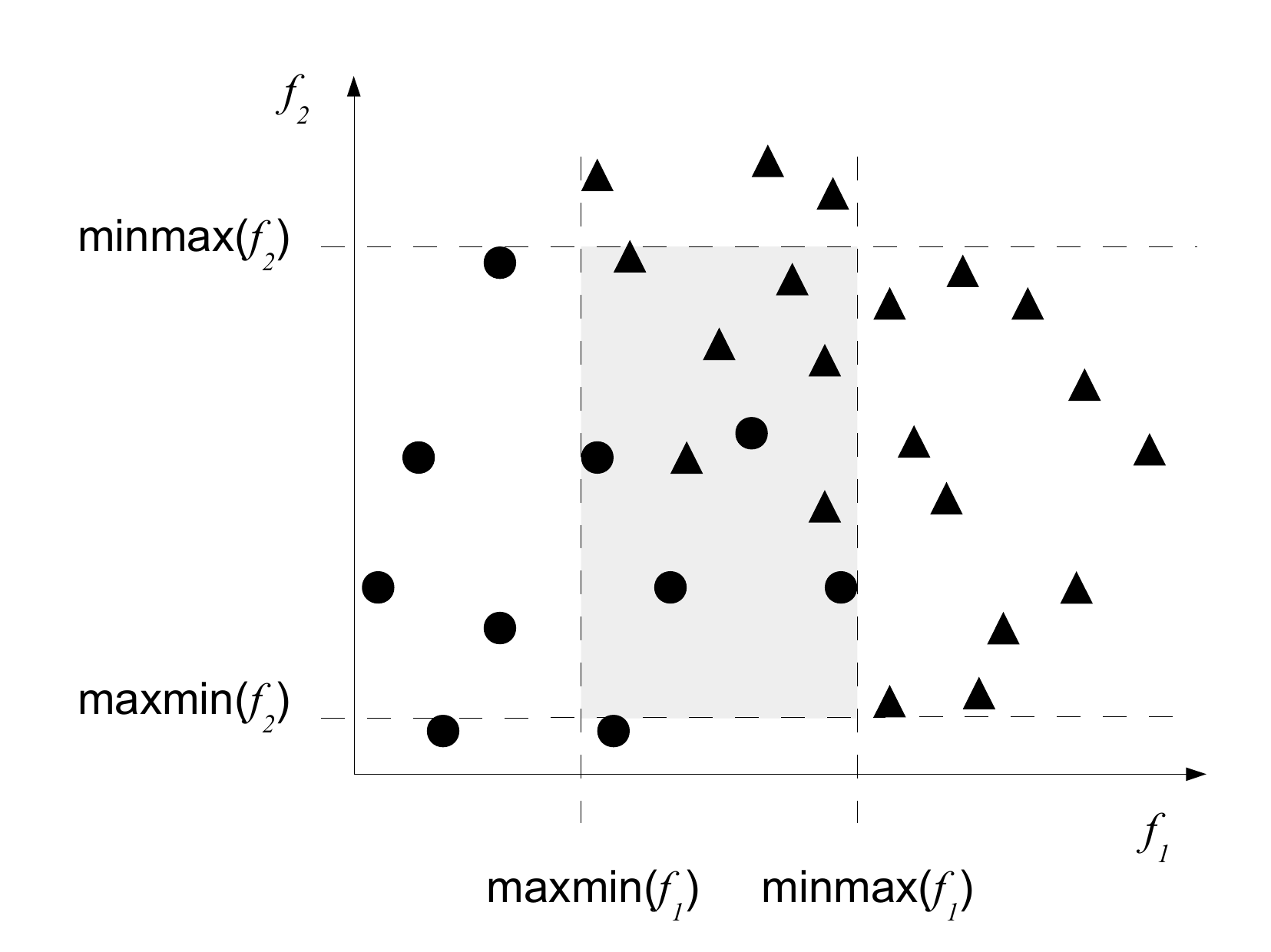}
	\caption{Example of overlapping region.}
	\label{fig:F2}
\end{figure}

The F2 measure calculates the overlap of the distributions of the features values within the classes. It can be determined by finding, for each feature $f_i$, its minimum and maximum values in the classes. The range of the overlapping interval is then calculated, normalized by the range of the values in both classes. Finally, the obtained values are multiplied, as shown in Equation \ref{eq:F2}.
\begin{equation} \label{eq:F2}
F2 = \prod_{i}^{m} \frac{overlap(f_i)}{range(f_i)} = \prod_{i}^{m} \frac{\max\{0,\min\max(f_i) - \max\min(f_i)\}}{\max\max(f_i) - \min\min(f_i)},
\end{equation}
where: 
\begin{align*}
& \min\max(f_i) = \min(\max(f_i^{c_1}),\max(f_i^{c_2})),
\\ 
& \max\min(f_i) = \max(\min(f_i^{c_1}),\min(f_i^{c_2})), 
\\ 
& \max\max(f_i) = \max(\max(f_i^{c_1}),\max(f_i^{c_2})), \\
& \min\min(f_i) = \min(\min(f_i^{c_1}),\min(f_i^{c_2})).
\end{align*}
The values $\max(f_i^{c_j})$ and $\min(f_i^{c_j})$ are the maximum and minimum values of each feature in a class $c_j \in \{1,2\}$, respectively. The numerator becomes zero when the per-class value ranges are disjoint for at least one feature. This equation uses a correction that was made in \citet{SoutoEtAl2010} and \citet{Cummins2013} to the original definition of F2, which may yield negative values for non-overlapping feature ranges. The asymptotic cost of this measure is $O(m \cdot n \cdot n_c)$, considering a OVO decomposition in the case of multiclass problems. The higher the F2 value, the greater the amount of overlap between the problem classes. Therefore, the problem's complexity is also higher. Moreover, if there is at least one non-overlapping feature, the F2 value should be zero. Figure \ref{fig:F2} illustrates the region that F2 tries to capture (as the shaded area), for a dataset with two features and two classes. 

\begin{figure}[ht!]
   \centering
   \subfloat[]{
        \label{fig:probF2c}         
        \includegraphics[scale = 0.2, clip, trim={0 8cm 0 6cm}]{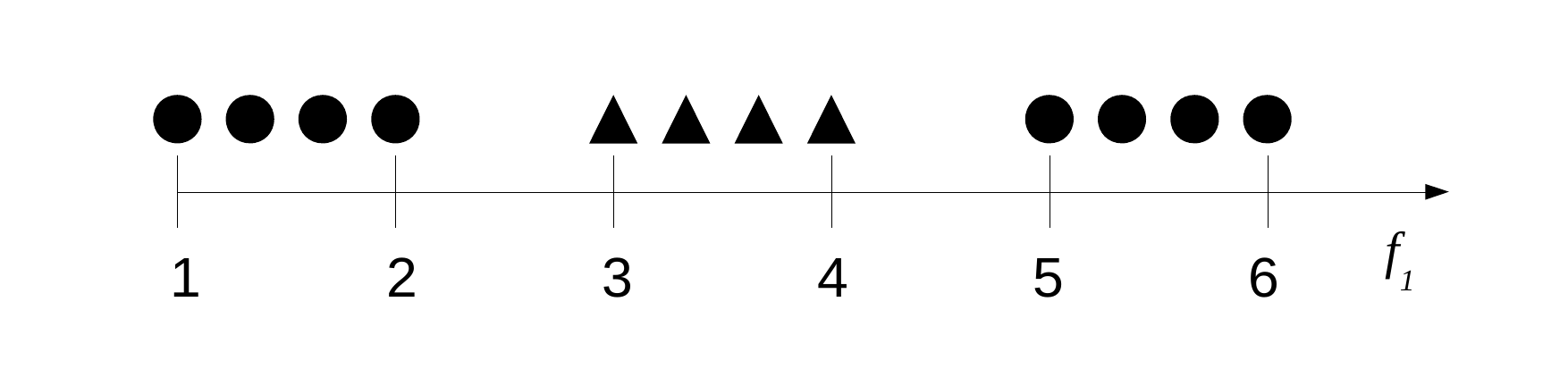}}
   \subfloat[]{
        \label{fig:probF2d}       
        \includegraphics[scale = 0.2, clip, trim={0 8cm 0 6cm}]{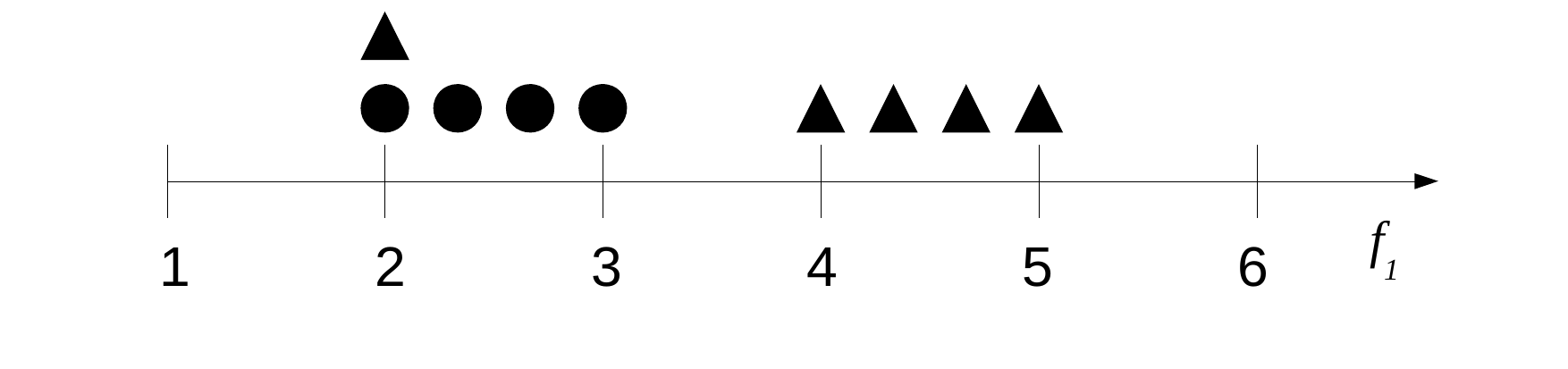}}
   \caption{Problematic situations for F2.}
   \label{fig:prob}               
\end{figure}

\citet{Cummins2013} points to an issue with F2 for the cases illustrated in Figure \ref{fig:prob}.  In Figure \ref{fig:probF2c}, the attribute is discriminative but the minimum and maximum values overlap in the different classes; and in Figure \ref{fig:probF2d}, there is one noisy example which disrupts the measure values. \citet{Cummins2013} proposes 
 to deal with 
 these situations  
 by counting the number of feature values in which there is an overlap, which is only suitable for discrete-valued features. Using this solution, continuous features must be discretized a priori, which imposes the difficulty of choosing a proper discretization technique and associated parameters, an open issue in data mining  \citep{kotsiantis2006discretization}. 

It should be noted that the situations shown in Figure \ref{fig:prob} can be also harmful for the F1 measure. 
As noted by \citet{hu2010selecting}, F2 does not capture the simplicity of a linear oblique border either, since it assumes again that the linear boundary is perpendicular to one of the features axes. Finally, the F2 value can become very small depending on the number of operands in Equation \ref{eq:F2}. That is, it is highly dependent on the number of features a dataset has. This worsens for problems with many features, so that their F2 values may not be comparable to those of other problems with fewer features. \citet{SoutoEtAl2010}, \citet{LorenaEtAl2012} and, more recently, \citet{seijo2019developing} use a sum instead of the product in Equation \ref{eq:F2}, which partially solves the problems identified. Nonetheless, the result is not an overlapping volume and corresponds to the amount or size of the overlapping region. In addition, the measure remains influenced by the number of features the dataset has.

\subsubsection{Maximum Individual Feature Efficiency (F3)} 

This measure estimates the individual efficiency of each feature in separating the classes, and considers the maximum value found among the $m$ features. Here we take the complement of this measure so that higher values are obtained for more complex problems. For each feature, it checks whether there is overlap of values between examples of different classes. If there is overlap, the classes are considered to be ambiguous in this region. The problem can be considered simpler if there is at least one feature which shows low ambiguity between the classes, so F3 can be expressed as: 

\begin{equation} 
\label{eq:F3}
F3 = \min_{i=1}^{m}\frac{n_{o}(f_i)}{n},
\end{equation}
where $n_{o}(f_i)$ gives the number of examples that are in the overlapping region for feature $f_i$ and can be expressed by Equation \ref{eq:po}. Low values of F3, computed by Equation \ref{eq:F3}, indicate simpler problems, where few examples overlap in at least one dimension. As with F2, the asymptotic cost of the F3 measure is $O(m \cdot n \cdot n_c)$.

\begin{equation}
\label{eq:po}
n_{o}(f_i) = \sum_{j=1}^n{I(x_{ji} > \max\min (f_i) \wedge x_{ji} < \min\max (f_i))}
\end{equation}

In Equation \ref{eq:po}, $I$ is the indicator function, which returns 1 if its argument is true and 0 otherwise, while $\max\min(f_i)$ and $\min\max(f_i)$ are the same as defined for F2. 

\begin{figure}[ht!]
   \subfloat[]{
        \label{fig:F3a}
        \includegraphics[scale = 0.25]{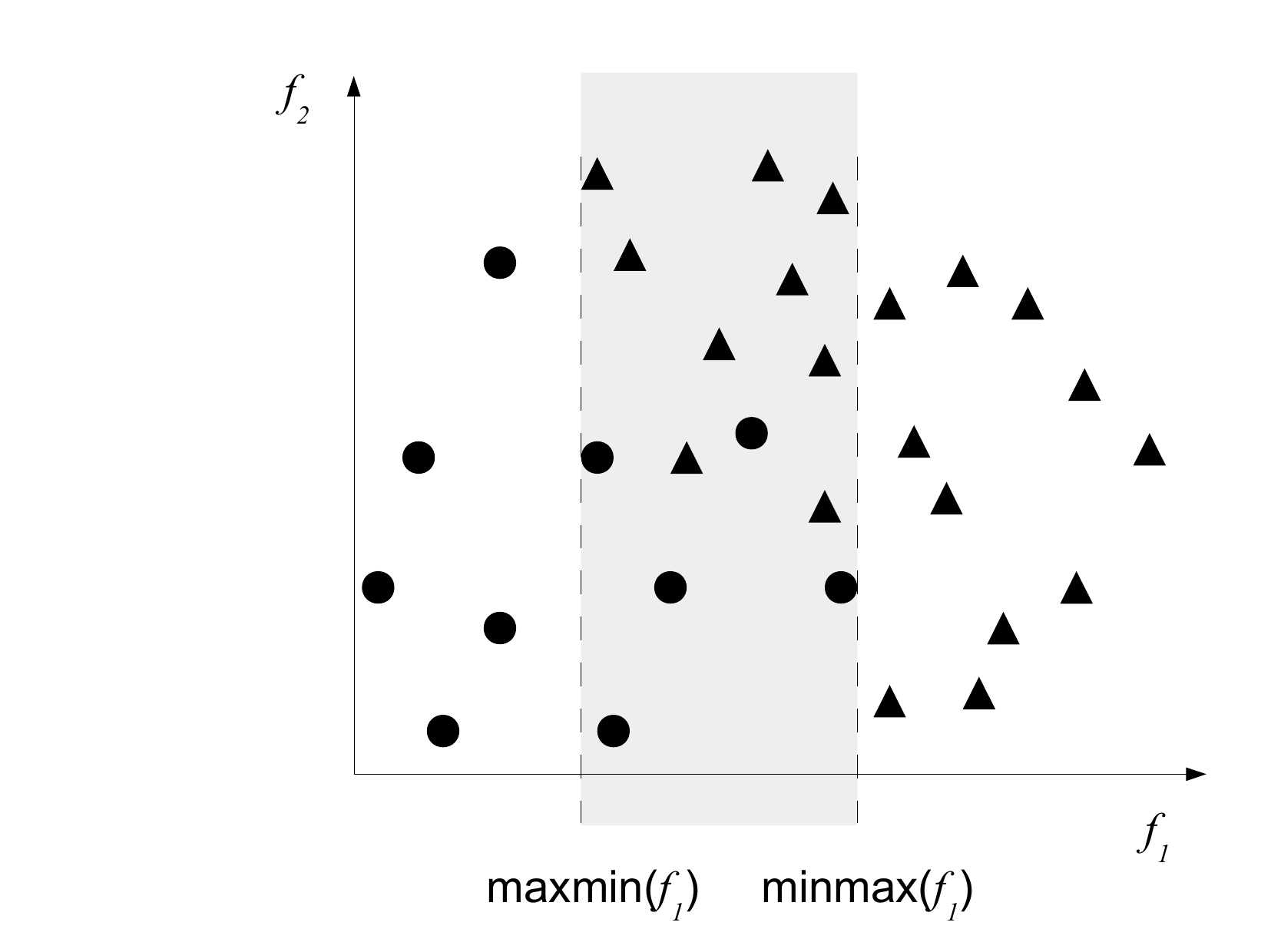}}
   \subfloat[]{
        \label{fig:F3b}
        \includegraphics[scale = 0.25]{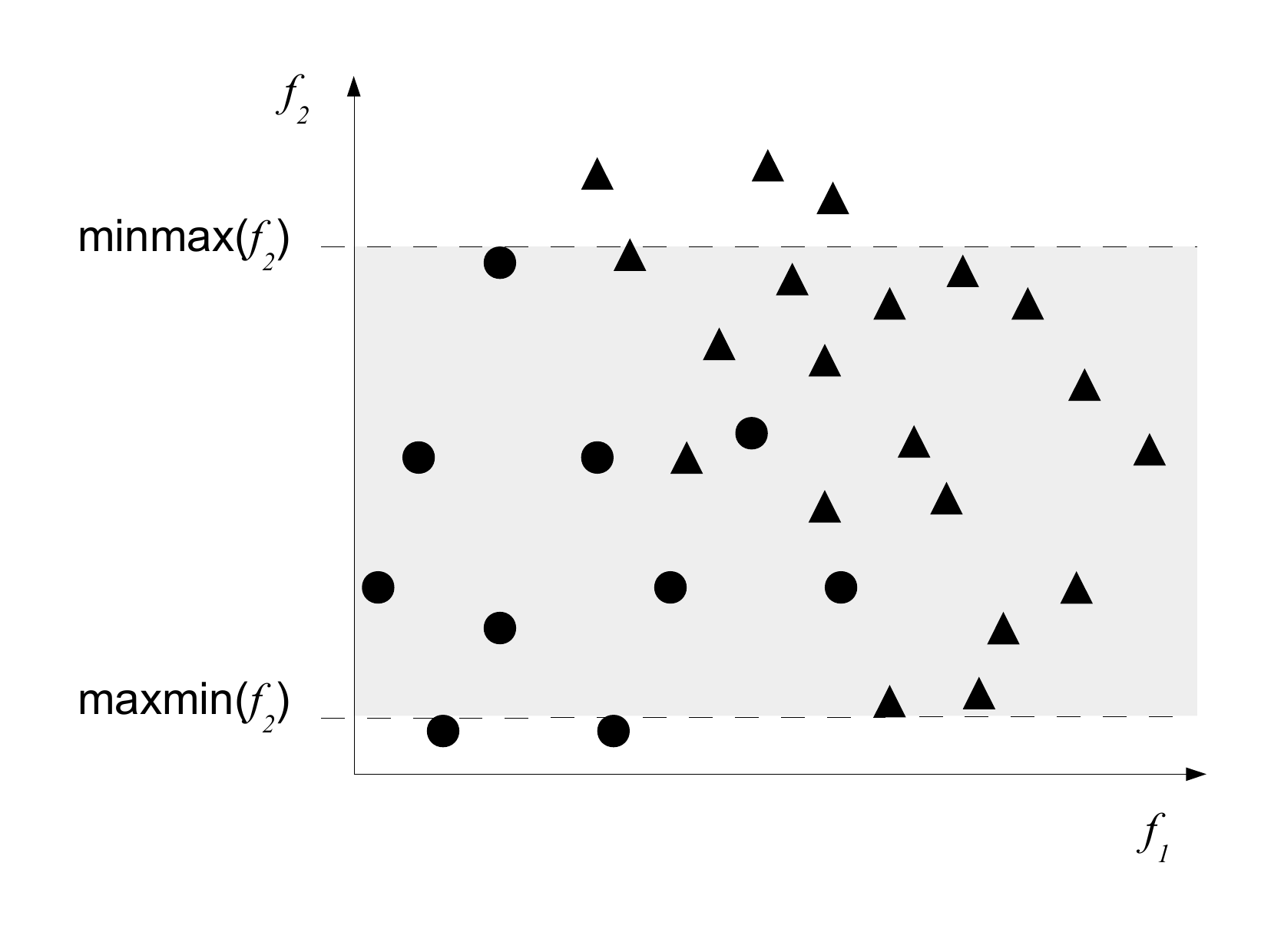}}
   \caption{Calculating F3 for the dataset from Figure \ref{fig:F2}.}\label{fig:F3}
\end{figure}

Figure \ref{fig:F3} depicts the computation of F3 for the same dataset from Figure \ref{fig:F2}. While for feature $f_1$ the proportion of examples that are  in the overlapping region is $\frac{14}{30}$ (Figure \ref{fig:F3a}), for $f_2$ this proportion is $\frac{25}{30}$  (Figure \ref{fig:F3b}), resulting in a F3 value of $\frac{14}{30}$. 

Since $n_{o}(f_i)$ is calculated taking into account the minimum and maximum values of the feature $f_i$ in different classes, it entails the same problems identified for F2 with respect to: classes in which the feature has more than one valid interval (Figure \ref{fig:probF2c}), susceptibility to noise (Figure \ref{fig:probF2d}) and the fact that it is assumed that in linearly separable problems, the boundary is perpendicular to an input axis.

\subsubsection{Collective Feature Efficiency (F4)} 
    
The F4 measure was proposed in \citet{Orriols-PuigEtAl2010} to get an overview of how the features work together. It successively applies a procedure similar to that adopted for F3. First the most discriminative feature according to F3 is selected; that is, the feature which shows less overlap between different classes. All examples that can be separated by this feature are removed from the dataset and the previous procedure is repeated: the next most discriminative feature is selected, excluding the examples already discriminated. This procedure is applied until all the features have been considered and can also be stopped when no example remains. F4 considers the ratio of examples that have not been discriminated, as presented in Equation \ref{eq:F4}. 
F4 is computed after $l$ rounds are performed through the dataset, where $l$  is in the range $[1,m]$. If one of the input features is already able to discriminate all the examples in $T$, $l$ is 1, whilst it can get up to $m$ in the case all features have to be considered. Its equation can be denoted by:   

\begin{equation}
\label{eq:F4}
F4 = \frac{n_o(f_{min}(T_l))}{n},
\end{equation}
where $n_{o}(f_{min}(T_l))$ measures the number of points in the overlapping region of feature $f_{min}$ for the dataset from the $l$-th round ($T_l$). This is the current most discriminative feature in $T_l$. Taking the $i$-th iteration of F4, the most discriminative feature in dataset $T_i$ can be found using Equation \ref{eq:feat}, adapted from F3.

\begin{equation}\label{eq:feat}
f_{min}(T_i) = \{f_j | \min_{j=1}^{m}{(n_{o}(f_j))}\}_{T_i}
\end{equation}
where $n_{o}(f_j)$ is computed according to Equation \ref{eq:po}. While the dataset at each round can be defined as: 

\begin{align}
&T_1 = T,\\
& T_i = T_{i-1} - \{\mathbf x_j | x_{ji} < \max\min (f_{min}(T_{i-1})) \vee x_{ji} > \min\max (f_{min}(T_{i-1}))
\end{align}

That is, the dataset at the $i$-th round is reduced by removing all examples that are already discriminated by the previous considered feature $f_{min}(T_{i-1})$. Therefore, the computation of F4 is similar to that of F3, except that it can be applied to reduced datasets. Lower values of F4 computed by Equation \ref{eq:F4} indicate that it is possible to discriminate more examples and, therefore, that the problem is simpler. The idea is to get the number
of examples that can be correctly classified if hyperplanes perpendicular to the axes of the features are used in their
separation. Since the overlapping measure applied is similar to that used for F3, they share the same problems in some estimates (as discussed for Figures \ref{fig:probF2c} and \ref{fig:probF2d}). F4 applies the F3 measure multiple times and at most it will iterate for all input features, resulting in a worst case asymptotic cost of $O(m^2 \cdot n \cdot n_c)$.

\begin{figure}[ht!]
   \centering
   \subfloat[]{
        \label{fig:F4a}
        \includegraphics[scale = 0.2, clip, trim={3.5cm 0cm 0 0cm}]{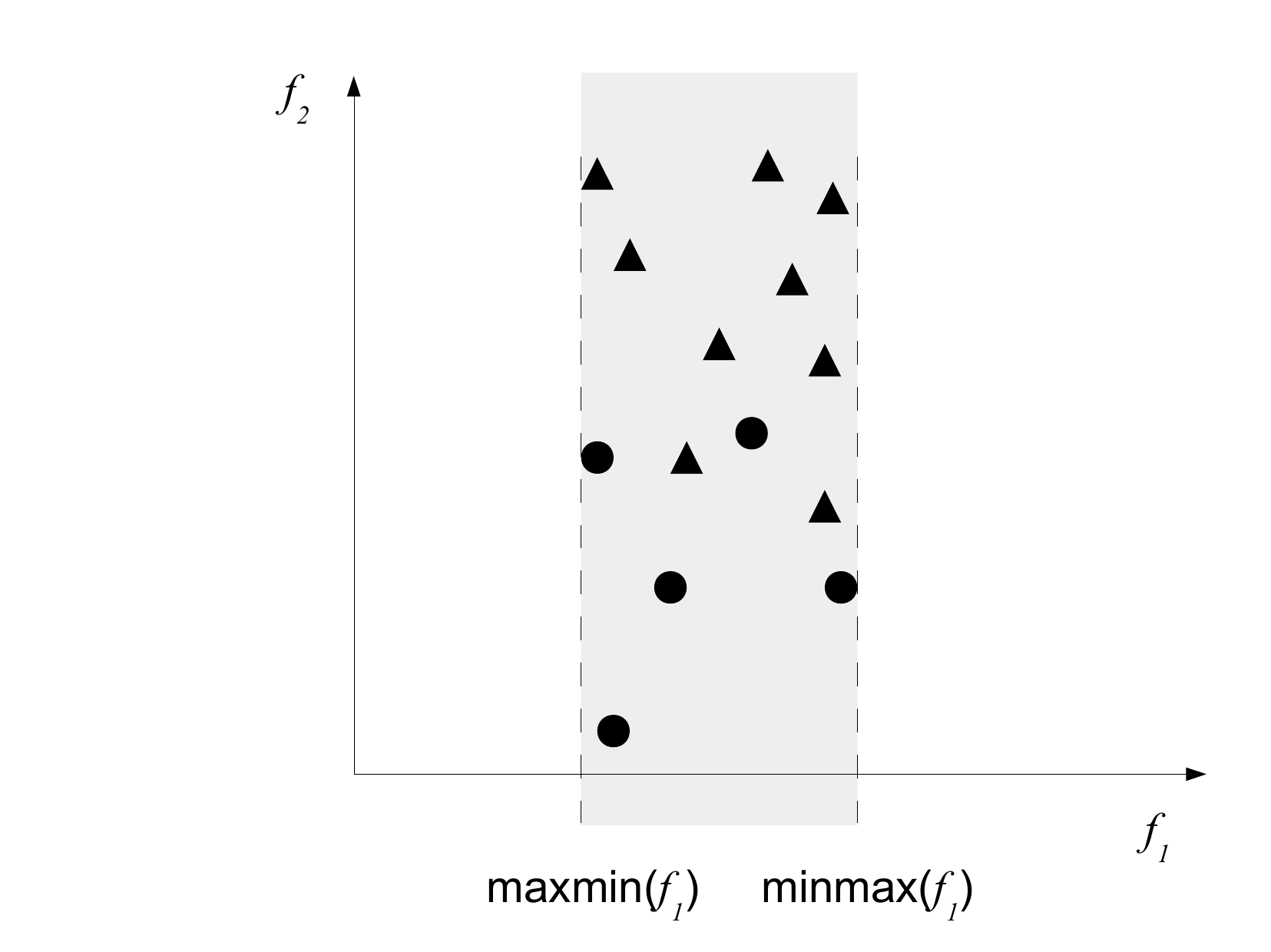}}
   \subfloat[]{
        \label{fig:F4b}
        \includegraphics[scale = 0.2, clip, trim={2cm 0cm 0 0cm}]{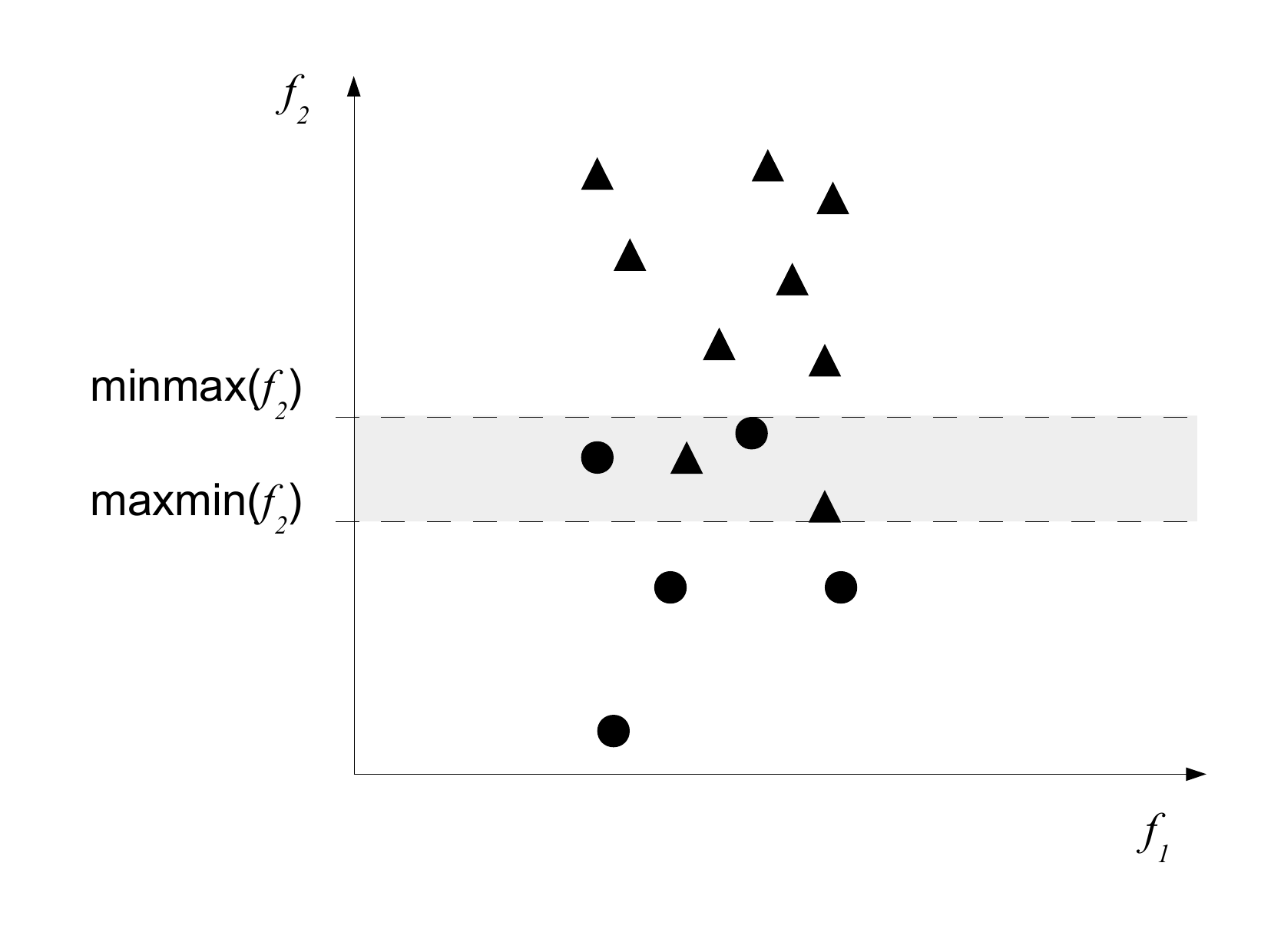}}
				\subfloat[]{
        \label{fig:F4c}
        \includegraphics[scale = 0.2, clip, trim={3cm 0cm 0 0cm}]{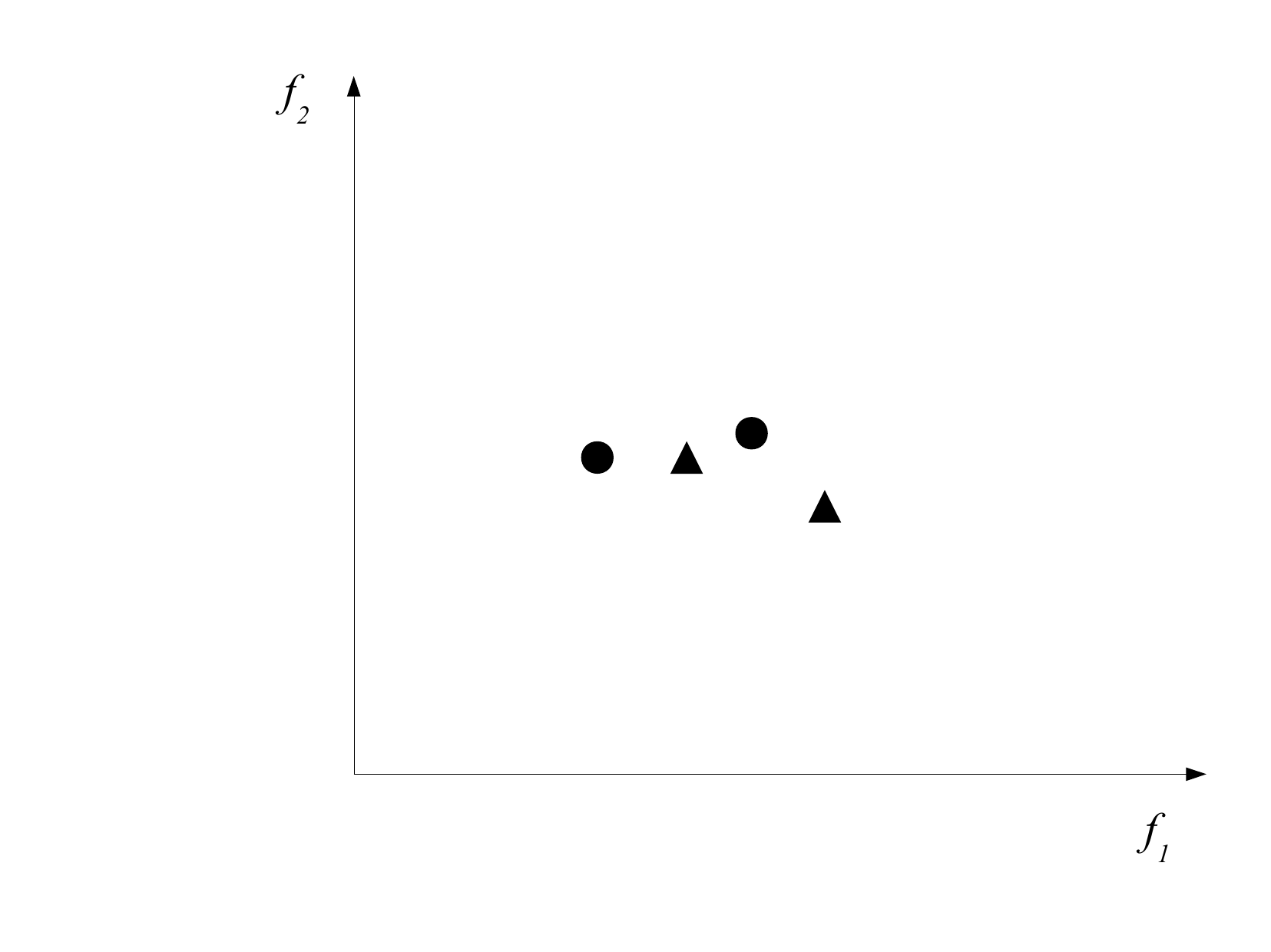}}
   \caption{Calculating F4 for the dataset from Figure \ref{fig:F2}.}
   \label{fig:F4}
\end{figure}

Figure \ref{fig:F4} shows the F4 operation for the dataset from Figure \ref{fig:F2}. Feature $f_1$ is the most discriminative in the first round (Figure \ref{fig:F3a}). Figure \ref{fig:F4a} shows the resulting dataset after all examples correctly discriminated by $f_1$ are disregarded. Figure \ref{fig:F4c} shows the final dataset after feature $f_2$ has been analyzed in Figure \ref{fig:F4b}. The F4 value for this dataset is $\frac{4}{30}$.

\subsection{Measures of Linearity}

These measures try to quantify to what extent the classes are linearly separable, that is, if it is possible to separate the classes by a hyperplane. They are motivated by the assumption that a linearly separable problem can be considered simpler than a problem requiring a non-linear decision boundary.  To obtain the linear classifier, \citet{HoBasu2002} suggest to solve an 
optimization problem proposed by \citet{smith1968pattern}, while in \citet{Orriols-PuigEtAl2010} a linear Support Vector Machine (SVM) \citep{cristianini2000introduction} is used instead. Here we adopt the SVM solution.
    
The hyperplane sought in the SVM formulation is the one which separates the examples from different classes with a maximum margin while minimizing training errors. This hyperplane is obtained by solving the following optimization problem: 
\begin{equation}\label{eq:funcao_objetivo_margens_suaves}
\mathop {\textnormal{Minimize}}\limits_{\mathbf{w},b,\mathbf \epsilon}
{\kern 1pt} {\kern 1pt} {\kern 1pt} {\kern 1pt} {\kern 1pt}
\frac{1} {2}\left\| \mathbf{w} \right\|^2 + C\left( \sum\limits_{i
= 1}^n {\varepsilon _i }\right)
\end{equation}
\begin{equation}\label{eq:restricoes_margens_suaves}
				\text{Subject to:} 
				\begin{cases}{\kern 3pt} y_i \left( {\mathbf{w} \cdot \mathbf{x}_i  + b} \right) \ge 1 - \varepsilon_i ,\\
        {\kern 5pt} \mathbf \varepsilon_i\ge 0, i = 1, \ldots,n
        \end{cases}
\end{equation}
where $C$ is the trade-off between the margin maximization, achieved by minimizing the norm of $\mathbf w$, and the minimization of the training errors, modeled by $\mathbf \varepsilon$. 
The hyperplane is given by $\mathbf w \cdot \mathbf x + b = 0$, where $\mathbf w$ is a weight vector and $b$ is an offset value. 
SVMs are originally proposed to solve binary classification problems 
with numerical features. Therefore, symbolic features must be converted into numerical values and multiclass problems must be first decomposed.

\subsubsection{Sum of the Error Distance by Linear Programming (L1)}

This measure
assesses if the data are linearly separable by computing, for a dataset, the sum of the distances of incorrectly classified examples to a linear boundary used in their classification. If the value of L1 is zero, then the problem is linearly separable and can be considered simpler than a problem for which a non-linear boundary is required.

Given the SVM hyperplane, the error distance of the erroneous instances can be computed by summing up the $\varepsilon_i$ values. 
For examples correctly classified with a margin larger than 1, $\varepsilon_i$ will be zero, whist it indicates the distance of the example to the linear boundary otherwise. This is expressed in Equation \ref{eq:SL1}.  
The $\varepsilon_i$ values are determined in the SVM optimization process. 

\begin{equation}\label{eq:SL1}
SumErrorDist = \frac{1}{n}\sum_{i=1}^{n} \varepsilon_i | h(\mathbf x_i) \neq y_i.
\end{equation}

The L1 value can then be computed as:
\begin{equation}\label{eq:L1}
L1 = 1 - \frac{1}{1+SumErrorDist} = \frac{SumErrorDist}{1+SumErrorDist}
\end{equation}

\begin{figure}[ht!]
   \centering
        \includegraphics[scale=0.2]
        {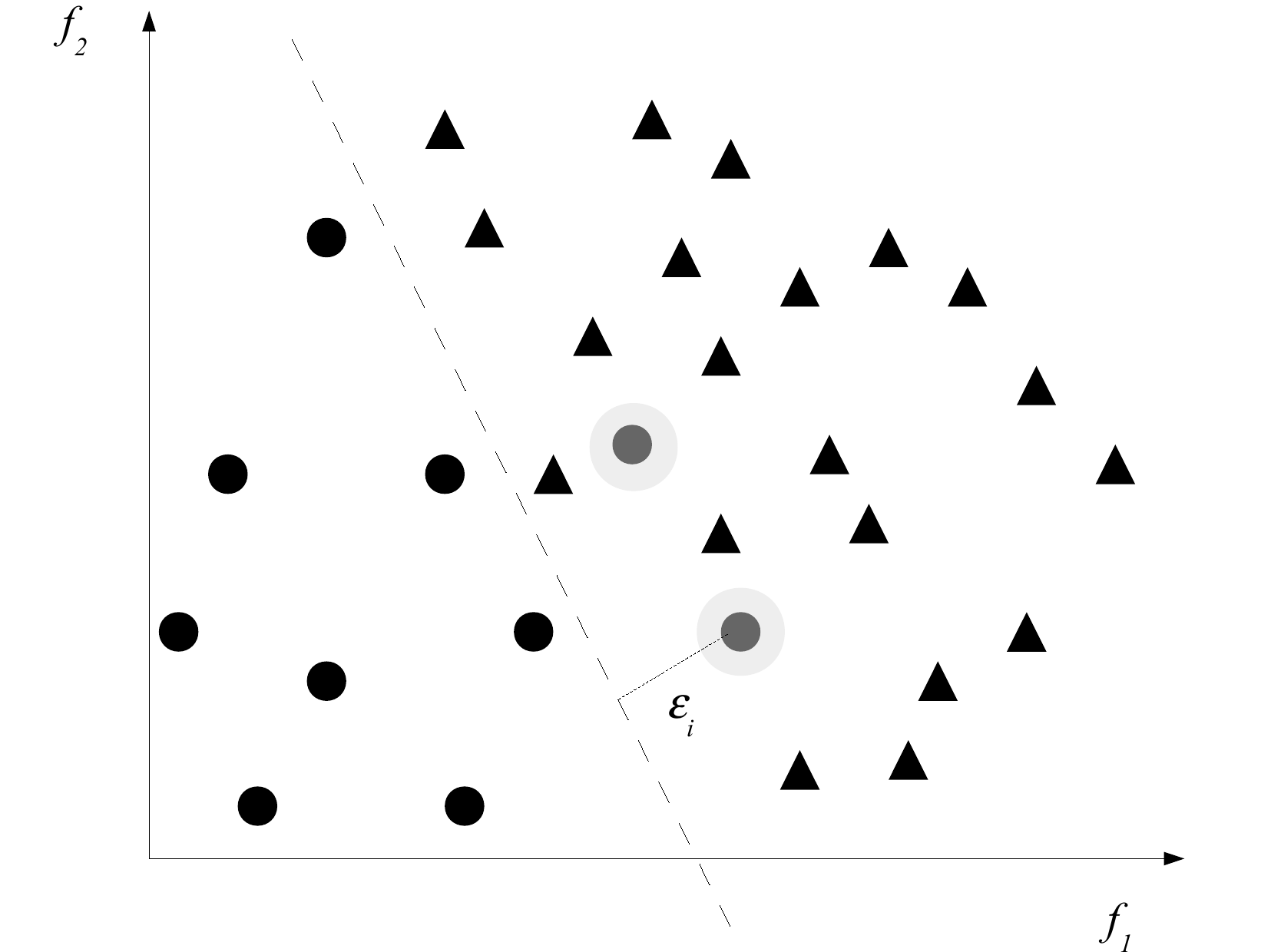}
   \caption{Example of L1 and L2 computation. The examples misclassified by the linear SVM are highlighted in gray.}
   \label{fig:L1}         
\end{figure}

Low values for L1 (bounded in $[0,1)$) indicate that the problem is close to being linearly separable, that is, simpler. 
Figure \ref{fig:L1} illustrates an example of L1 application. After a linear boundary is obtained, the $\epsilon_i$ values of the misclassified examples (gray circles) are summed up and subject to Equation \ref{eq:L1}.

L1 does not allow to check if a linearly separable problem is simpler than another that is also linearly separable. Therefore, a dataset for which data are distributed narrowly along the linear boundary will have a null L1 value, and so will a dataset in which the classes are far apart with a large margin of separation. The asymptotic computing cost of the measure is dependent on that of the linear SVM, and can take $O(n^2)$ operations in the worst case \citep{bottou2007support}. In multiclass classification problems decomposed according to OVO, this cost would be $O(n_c^2 \cdot (\frac{n}{n_c})^2)$, which resumes to $O(n^2)$ too. 

\subsubsection{Error Rate of Linear Classifier (L2)}

The L2 measure
computes the error rate of the linear SVM classifier. Let $h(\mathbf x)$ denote the linear classifier obtained. L2 is then given by:
\begin{equation}
\label{eq:L2}
L2 = \frac{\sum_{i=1}^{n}{I(h(\mathbf x_i) \neq y_i)}}{n}
\end{equation}

Higher L2 values denote more errors and therefore a greater complexity regarding the aspect that the data cannot be separated  linearly. For the dataset in Figure \ref{fig:L1}, the L2 value is $\frac{2}{30}$. L2 has similar issues with L1 in that it does not differentiate between problems that are barely linearly separable (i.e., with a narrow margin) from those with classes that are very far apart.  The asymptotic cost of L2 is the same of L1, that is, $O(n^2)$.  

\subsubsection{Non-Linearity of a Linear Classifier (L3)} 

\begin{figure}[ht!]
	\centering
	\includegraphics[scale=0.2]{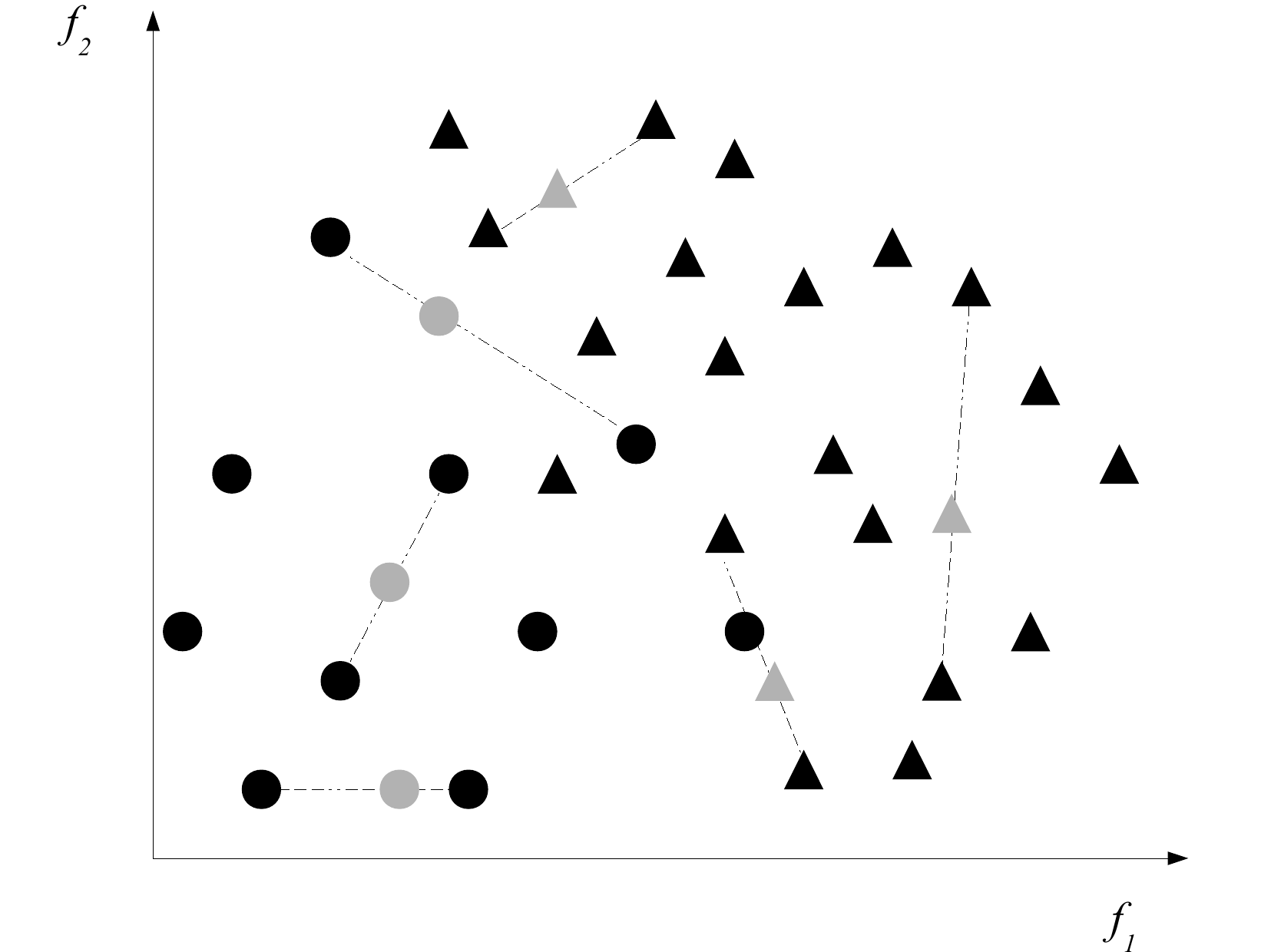}
	\caption{Example of how new points are generated in measures L3 and N4.}
	\label{fig:L3}
\end{figure}

This measure
uses a methodology proposed by \citet{hoekstra1996nonlinearity}. It first creates a new dataset by interpolating pairs of training examples of the same class. Herewith, two examples from the same class are chosen randomly and they are linearly interpolated (with random coefficients), producing a new example. Figure \ref{fig:L3} illustrates the generation of six new examples (in gray) from a base training dataset. 
Then a linear classifier 
is trained on the original data and has its error rate measured in the new data points. 
This index is sensitive to how the data from a class are distributed in the border regions and also on how much the convex hulls which delimit the classes overlap. 
In particular, it detects the presence of concavities in the class boundaries \citep{armano2016experimenting}. Higher values indicate a greater complexity.
Letting $h_T(\mathbf x)$ denote the linear classifier induced from the original training data $T$, the L3 measure can be expressed by:
\begin{equation}
\label{eq:L3}
L3 = \frac{1}{l}\sum_{i=1}^{l}{I(h_T(\mathbf x_i') \neq y_i')},
\end{equation}
where $l$ is the number of interpolated examples $\mathbf x_i'$ and their corresponding labels are denoted by $y_i'$.  
In ECoL we generate the interpolated examples maintaining the proportion of examples per class from the original dataset and use $l = n$. The asymptotic cost of this measure is dependent on both the induction of a linear SVM and the time taken to obtain the predictions for the $l$ test examples, resulting in $O(n^2 + m \cdot l \cdot n_c)$.

\subsection{Neighborhood Measures}

These measures try to capture the shape of the decision boundary and characterize the class overlap by analyzing local neighborhoods of the data points. Some of them also capture the internal structure of the classes. All of them work over a distance matrix storing the distances between all pairs of points in the dataset. To deal with both symbolic and numerical features, we adopt a heterogeneous distance measure named Gower \citep{gower1971general}. For symbolic features, the Gower metric computes if the compared values are equal, whilst for numerical features, a normalized difference of values is taken.

\subsubsection{Fraction of Borderline Points (N1)}

\begin{figure}[ht!]
	\centering
	\includegraphics[width=0.35\textwidth]{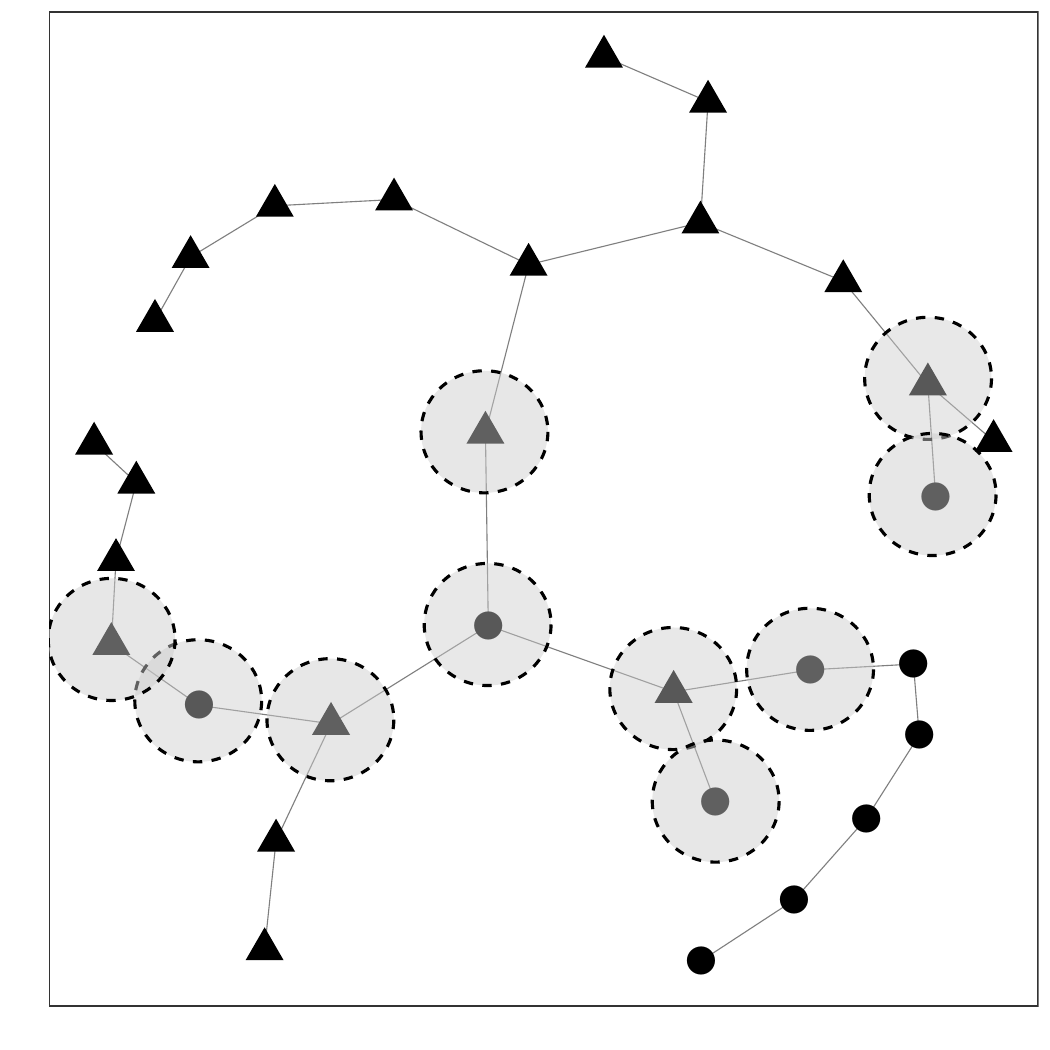}
	\caption{Example of MST generated for the dataset from Figure \ref{fig:F2} and the detected points in the decision border.}
	\label{fig:AGM}
\end{figure}

In this measure, 
first a Minimum Spanning Tree (MST) is built from data, as illustrated in Figure \ref{fig:AGM}. 
Herewith, each vertex corresponds to an example and the edges are weighted according to the distance between them. N1 is obtained by computing the percentage of vertices incident to edges connecting examples of opposite classes in the generated MST. These examples are either on the border or in overlapping areas between the classes. They can also be noisy examples surrounded by examples from another class. Therefore, N1 estimates the size and complexity of the required decision boundary through the identification of the critical points in the dataset: those very close to each other but belong to different classes. Higher N1 values indicate the need for more complex boundaries to separate the classes and/or that there is a large amount of overlapping between the classes. N1 can be expressed as:
\begin{equation}
\label{eq:N1}
N1 = \frac{1}{n}\sum_{i=1}^{n}{I((\mathbf x_i,\mathbf x_j)  \in MST {\kern 3pt} \wedge {\kern 3pt} y_i \neq y_j) }
\end{equation}

To build the graph from the data, it is necessary to first compute the distance matrix between all pairs of elements, which requires $O(m \cdot n^2)$ operations. Next, using \emph{Prim's algorithm} for obtaining the MST requires $O(n^2)$ operations in the worst case. Therefore, the total asymptotic complexity of N1 is $O(m \cdot n^2)$.

N1 is sensitive to the type of noise where the closest neighbors of noisy examples have a different class from their own, as typical in the scenario where erroneous class labels are introduced during data preparation. Datasets with this type of noise are considered more complex than their clean counterparts, according to the N1 measure, as observed in 
\citet{LorenaEtAl2012} and \citet{GarciaEtAl2015}. 

Another issue 
is that there can be multiple MSTs valid for the same set of points. 
\citet{Cummins2013} propose to generate ten MSTs by presenting the data points in different orderings and reporting an average N1 value. \citet{BasuHo2006} also report that the N1 value can be large even for a linearly separable problem. This happens when the distances between borderline examples are smaller than the distances between examples from the same class. 
On the other hand, \citet{Ho2002} suggests that a problem with a complicated nonlinear class boundary can still have relatively few edges among examples from different classes as long as the data points are compact within each class. 

\subsubsection{Ratio of Intra/Extra Class Nearest Neighbor Distance (N2)}

\begin{figure}[ht!]
	\centering
	\includegraphics[width=0.4\textwidth]{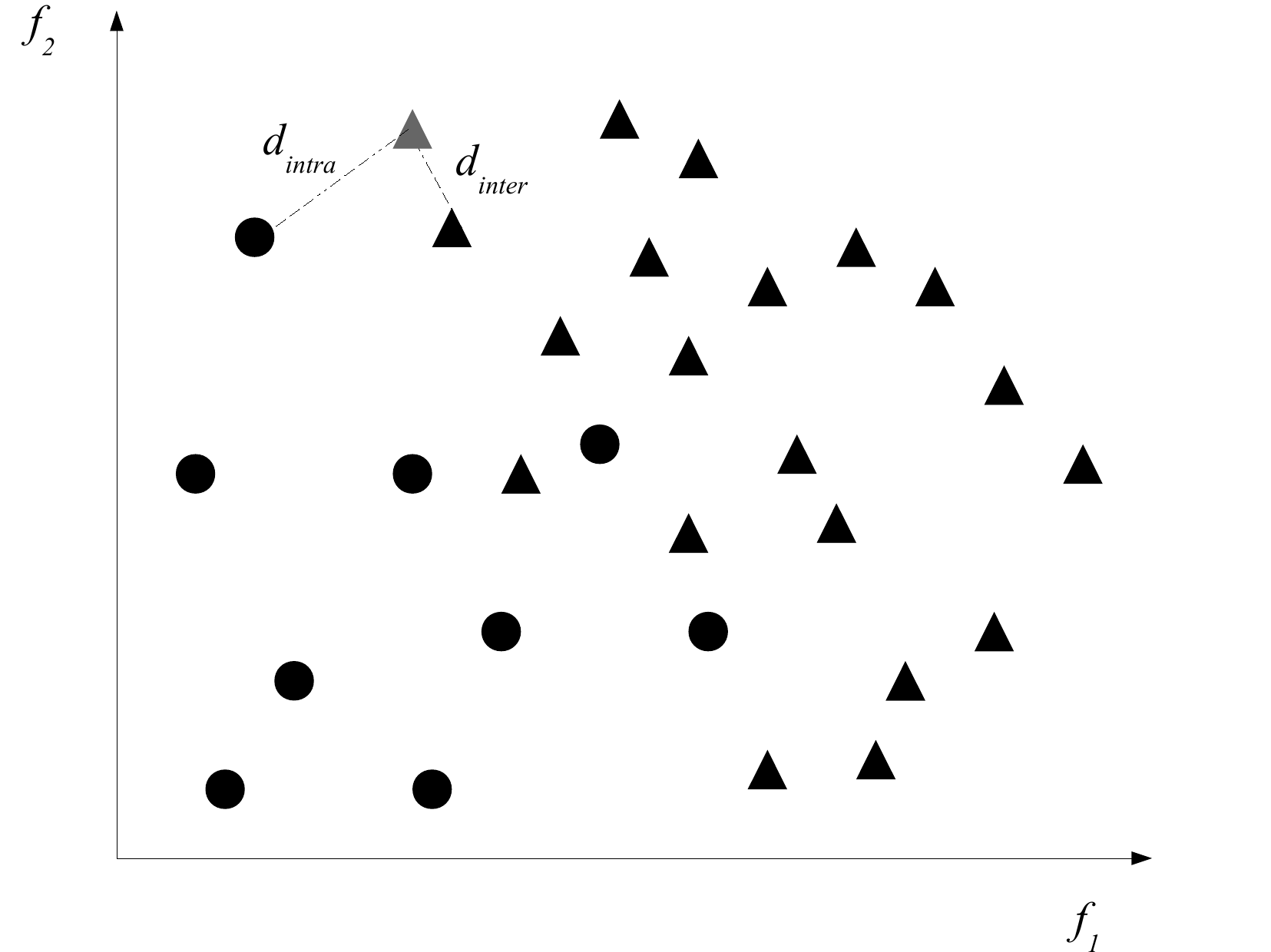}
	\caption{Example of intra and inter class distances for a particular example.}
	\label{fig:N2}
\end{figure}

This measure 
computes the ratio of two sums: (i) the sum of the distances between each example and its closest neighbor from the same class (intra-class); and (ii) the sum of the distances between each example and its closest neighbor from another class (extra-class). This is shown in Equation \ref{eq:N2}.
    \begin{equation}\label{eq:IN2}
        intra\_extra = \frac{\sum_{i=1}^{n}{d(\mathbf{x_i},NN(\mathbf x_i) \in y_i)}}{\sum_{i=1}^{n}{d(\mathbf{x_i},NN(\mathbf x_i) \in y_j \neq y_i)}},
    \end{equation}
where $d(\mathbf{x_i},NN(\mathbf x_i) \in {y_i})$ corresponds to the distance of example $\mathbf x_i$ to its nearest neighbor (NN) from its own class $y_i$ and $d(\mathbf{x_i},NN(\mathbf x_i) \in {y_j \neq y_i})$ represents the distance of $\mathbf x_i$ to the closest neighbor from another class $y_j \neq y_i$ ($\mathbf x_i$'s nearest enemy). Based on the intra/extra class calculation, N2 can be obtained as:
\begin{equation}\label{eq:N2}
        N2 = 1 - \frac{1}{1+intra\_extra} = \frac{intra\_extra}{1+intra\_extra}.
    \end{equation}
The computation of N2 requires obtaining the distance matrix between all pairs of elements in the dataset, which requires $O(m \cdot n^2)$ operations. Figure \ref{fig:N2} illustrates the intra- and extra-class distances for a particular example in a dataset. 

Low N2 values are indicative of simpler problems, in which the overall distance between examples of different classes exceeds the overall distance between examples from the same class. N2 is sensitive to how data are distributed within classes and not only to how the boundary between the classes is like. It can also be sensitive to labeling noise in the data, just like N1. 
According to \citet{Ho2002}, a high N2 value can also be obtained for a linearly separable problem where the classes are distributed in a long, thin, and sparse structure along the boundary. 
It must be also observed that N2 is related to F1 and F1v, since they all assess intra and inter class variabilities.  However, 
N2 uses a distance that summarizes the joint relationship between the values of all the features for the concerned examples. 

\subsubsection{Error Rate of the Nearest Neighbor Classifier (N3)}	

The N3 measure refers to the error rate of a 1NN 
classifier that is estimated using a leave-one-out procedure. The following equation denotes this measure:
\begin{equation}
\label{eq:N3}
N3 = \frac{\sum_{i=1}^{n}{I(NN(\mathbf x_i) \neq y_i)}}{n},
\end{equation}
where $NN(\mathbf x_i)$ represents the nearest neighbor classifier's prediction for example $\mathbf x_i$ using all the others as training points. 
High N3 values indicate that many examples are close to examples of other classes, making the problem more complex. N3 requires $O(m \cdot n^2)$ operations.

\subsubsection{Non-Linearity of the Nearest Neighbor Classifier (N4)} 

This measure 
is similar to L3, but uses the NN classifier instead of the linear predictor. It can be expressed as:
\begin{equation}
\label{eq:N4}
N4 = \frac{1}{l}\sum_{i=1}^{l}{I(NN_T(\mathbf x_i') \neq y_i')},
\end{equation}
where $l$ is the number of interpolated points, generated as illustrated in Figure \ref{fig:L3}. 
Higher N4 values are indicative of problems of greater complexity. 
In contrast to L3, N4 can be applied directly to multiclass classification problems, without the need to decompose them into binary subproblems first. The asymptotic cost of computing N4 is $O(m \cdot n \cdot l)$ operations, as it is necessary to compute the distances between all possible testing and training examples.

\subsubsection{Fraction of Hyperspheres Covering Data (T1)}

This is regarded as a topological measure in \citet{HoBasu2002}.  It uses a process that builds hyperspheres centered at each one of the examples. The radius of each hypersphere is progressively increased until the hypersphere reaches an example of another class. Smaller hyperspheres contained in larger hyperspheres are eliminated. T1 is defined as the ratio between the number of the remaining hyperspheres and the total number of examples in the dataset:
\begin{equation}
\label{eq:T1}
T1 = \frac{\sharp Hyperspheres(T)}{n}
\end{equation}
where $\sharp Hyperspheres(T)$ gives the number of hyperspheres that are needed to cover the dataset. 

The hyperspheres represent a form of 
adherence subsets as discussed in \citet{lebourgeois1996}.
The idea is to obtain an adherence subset of maximum order for each example 
such that it includes only examples from the same class. Subsets that are completely included in other subsets are discarded. In principle the adherence subsets can be of any form (e.g. hyperectangular), and hyperspheres are chosen in the definition of this measure because it can be defined with relatively few parameters (i.e., only a center and a radius).
Fewer hyperspheres are obtained for simpler datasets. This happens when data from the same class are densely distributed and close together. Herewith, this measure also captures the distribution of data within the classes and not only their distribution near the class boundary.

\begin{figure}[ht!]
   \centering
   \subfloat[Hyperspheric radiuses of two examples that are mutual nearest enemies]{
        \label{fig:T1_1}
        \includegraphics[scale=0.2, clip, trim={2.0cm 0cm 2.5cm 0cm}]{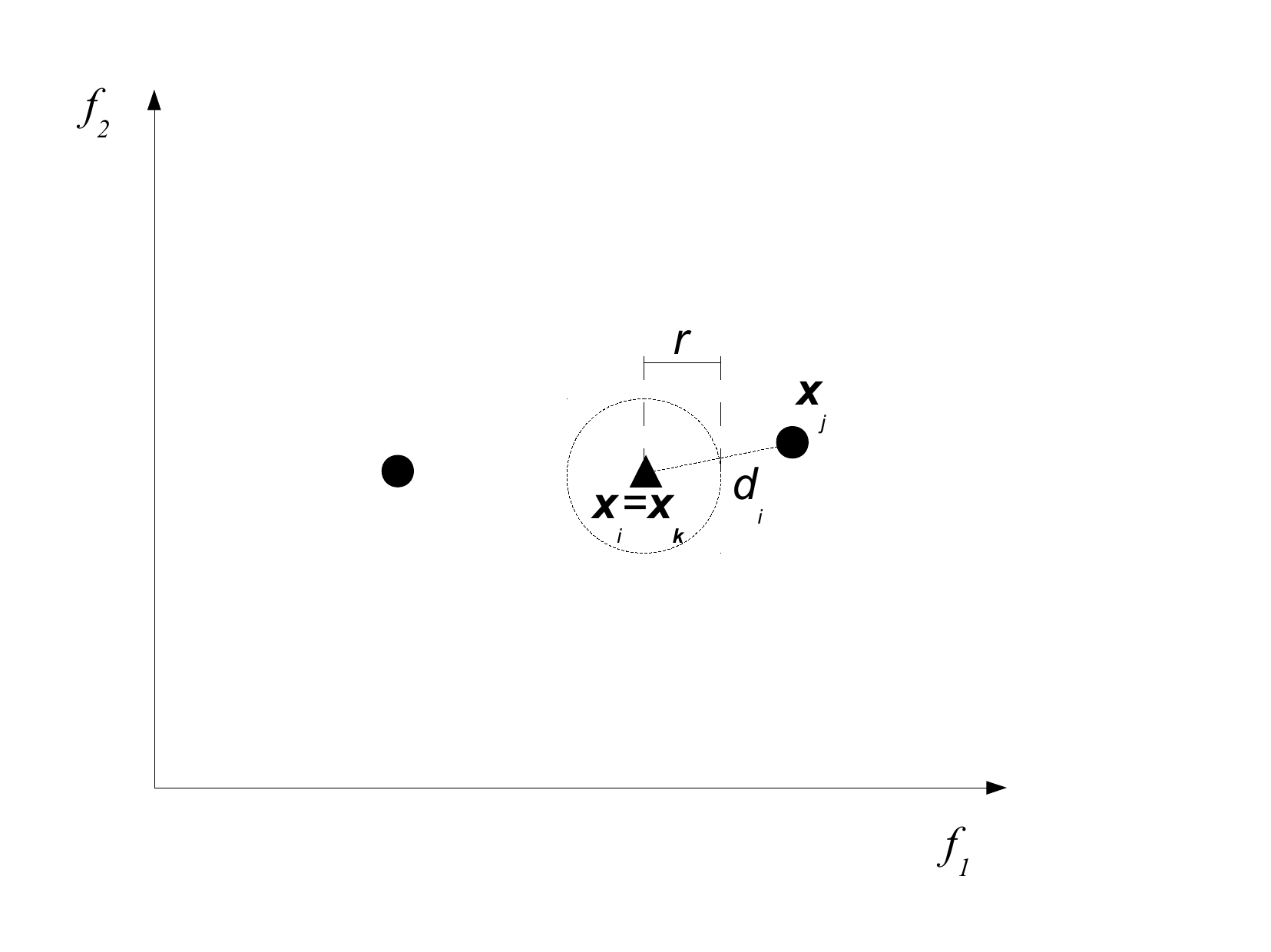}}
   \subfloat[Radius $r$ obtained recursively]{
        \label{fig:T1_2}
        \includegraphics[scale=0.2, clip, trim={2.0cm 0cm 2.5cm 0cm}]{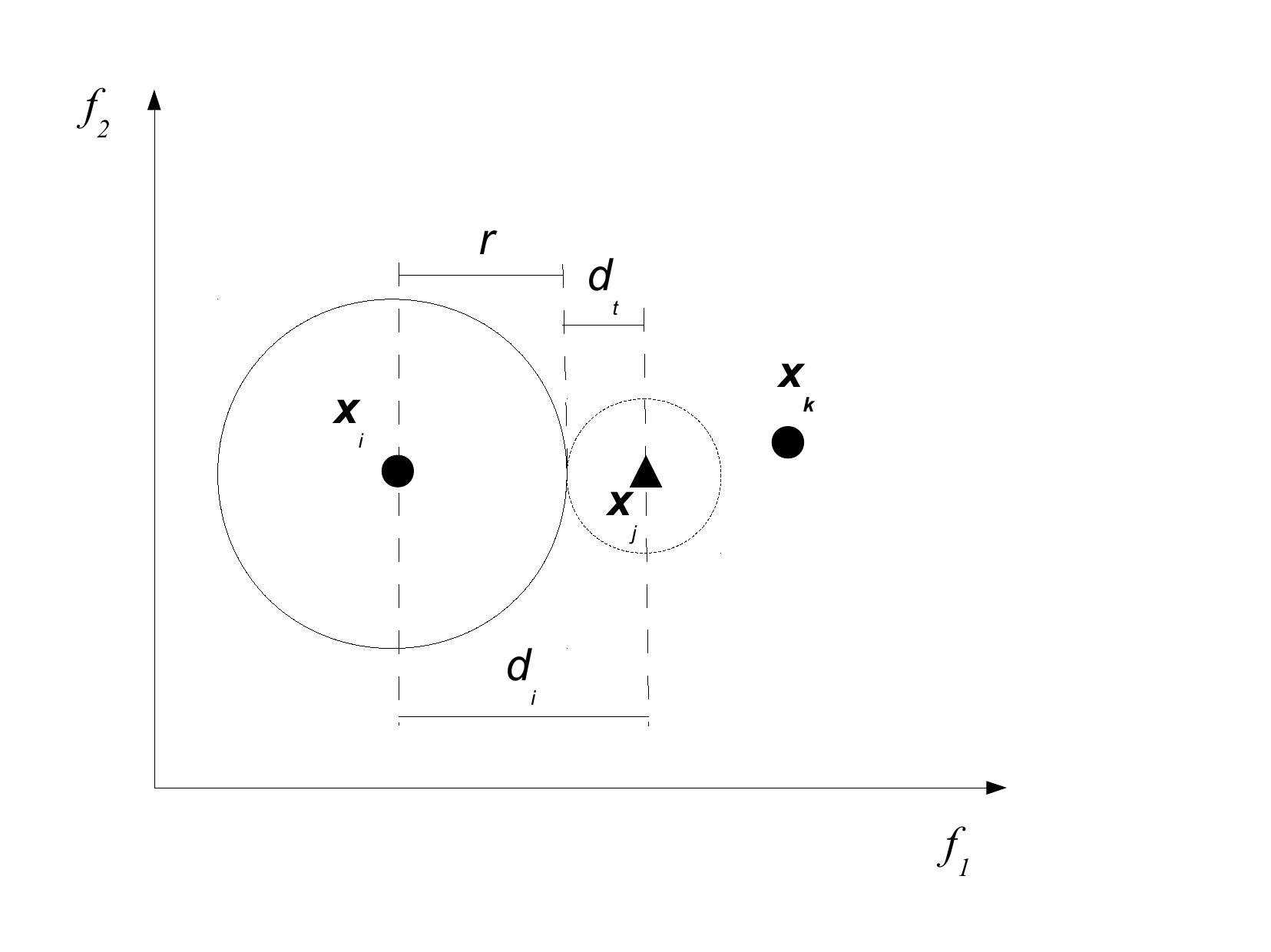}}	\subfloat[Final hyperspheres for a dataset]{
        \label{fig:T1_3}
        \includegraphics[scale=0.2, clip, trim={1.2cm 0cm 1.0cm 0cm}]{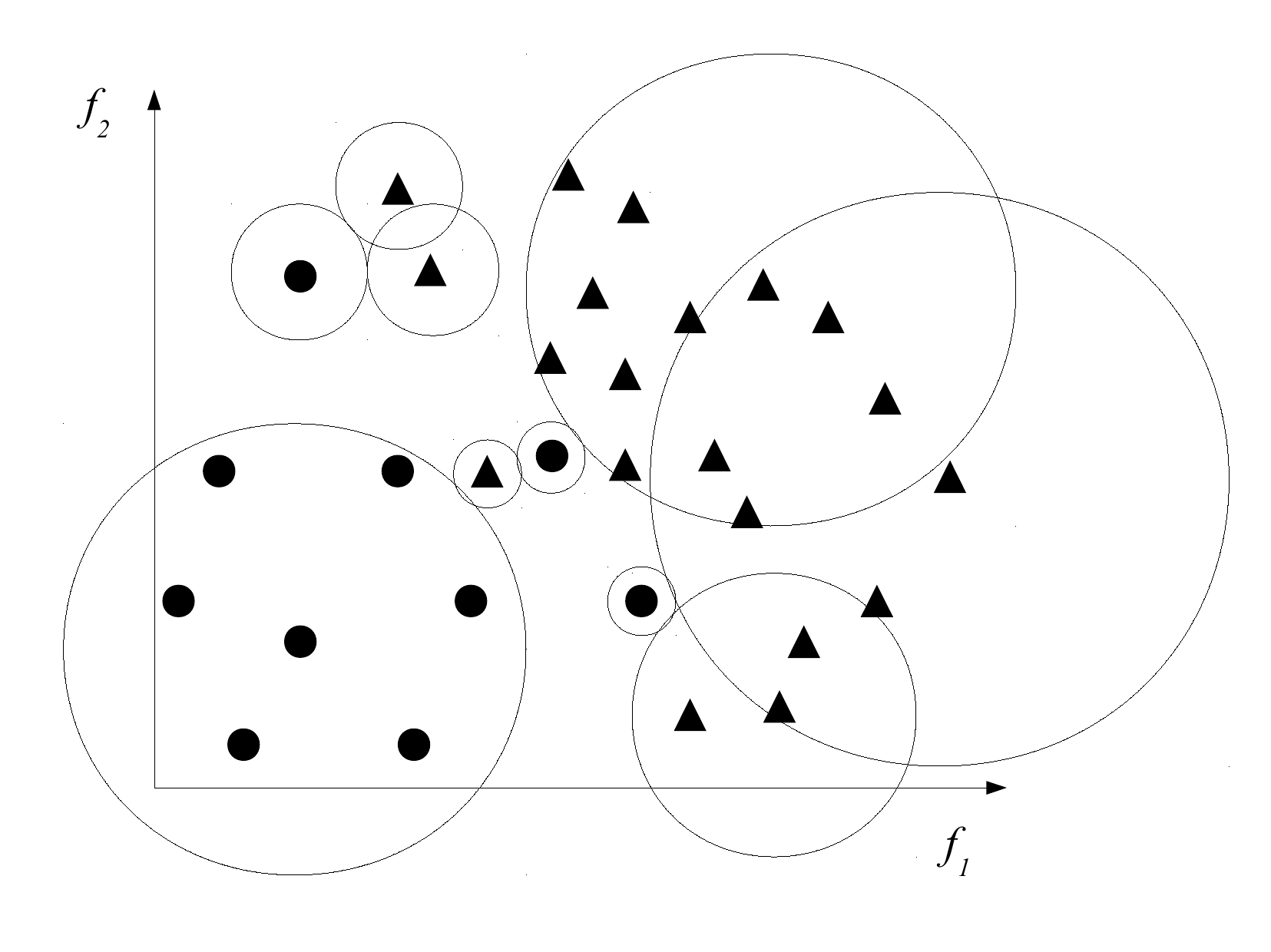}}
   \caption{Calculating T1 for a dataset.}
   \label{fig:T1}
\end{figure}

In this paper we propose an alternative implementation of T1. It involves a modification of the definition to stop the growth of the hypersphere when the hyperspheres centered at two points of opposite classes just start to touch.  With this modification, the radius of each hypersphere around an example can be directly determined based on distance matrix between all examples. 
The radius computation for an example $\mathbf x_i$ is shown in Algorithm \ref{alg:radios}, in which the nearest enemy ($ne$) of an example corresponds to the nearest data point from an opposite class ($ne(\mathbf x_i) = NN(\mathbf x_i) \in y_j \neq y_i$). 
If two points are mutually nearest enemies of each other (line 3 in Algorithm \ref{alg:radios}), the radiuses of their hyperspheres correspond to half of the distance between them (lines 4 and 5, also see Figure \ref{fig:T1_1}). The radiuses of the hyperspheres around other examples can be determined recursively (lines 7 and 8), as illustrated in Figure \ref{fig:T1_2}. 

Once the radiuses of all hyperspheres are found, a post-processing step can be applied to verify which hyperspheres are absorbed: 
those lying inside larger hyperspheres. The hyperspheres obtained for our example dataset is shown in Figure \ref{fig:T1_3}. 
The most demanding operation in T1 is to compute the distance matrix between all the examples in the dataset, which requires $O(m \cdot n^2)$ operations.

\begin{algorithm}[h] \caption{Computing the radius of the hypersphere of an example $\mathbf x_i$.}\label{alg:radios}
\begin{algorithmic}[1]
\REQUIRE A distance matrix $D_{n \textnormal{x} n}$, a label vector $\mathbf y$, a data index $i$;\\
\STATE $\mathbf x_j$ = $ne(\mathbf x_i$); 
\STATE $d_i =$ distance of $\mathbf x_i$ to $\mathbf x_j$;
\STATE $\mathbf x_k$ = $ne(\mathbf x_j)$;
\IF{($\mathbf x_i = \mathbf x_k$)}
        \RETURN{$\frac{d_i}{2}$}; 
    \ELSE
\STATE $d_t$ = radius($D$, $\mathbf y$, $j$);\\
        \RETURN{$d_i - d_t$};
    \ENDIF
\end{algorithmic}
\end{algorithm}

\subsubsection{Local Set Average Cardinality (LSC)}

\begin{figure}[ht!]
   \centering
        \includegraphics[width=0.4\textwidth]
        {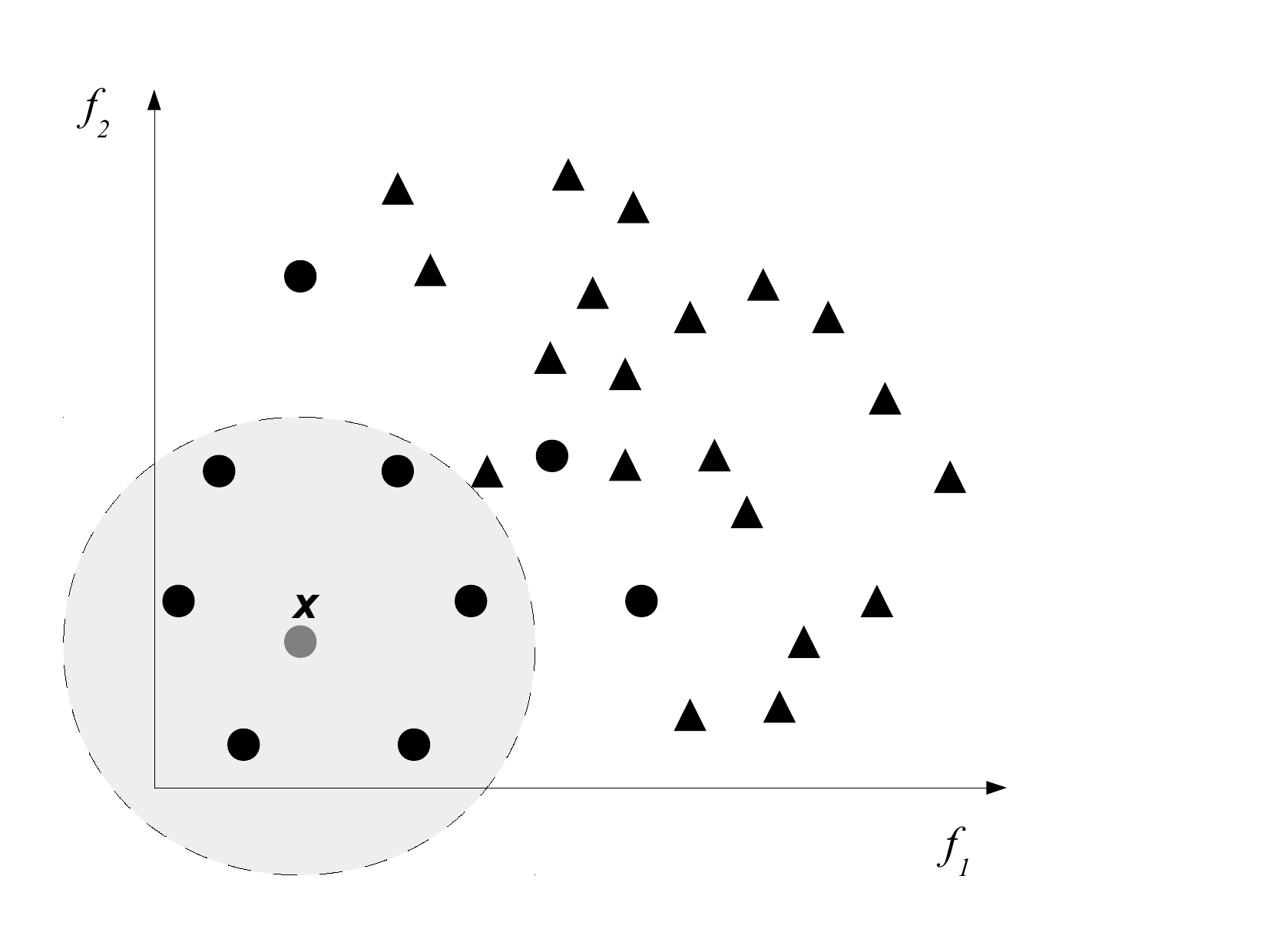}
   \caption{Local set of an example $\mathbf x$ in a dataset.}\label{fig:LSA}
\end{figure}

According to \citet{leyvaset}, the Local-Set (LS) of an example $\mathbf x_i$ in a dataset ($T$) is defined as the set of points from $T$ whose distance to $\mathbf x_i$ is smaller than the distance from $\mathbf x_i$ to $\mathbf x_i$'s nearest enemy (Equation \ref{eq:LS}). 
\begin{equation}\label{eq:LS}
LS(\mathbf x_i) = \{\mathbf x_j | d(\mathbf x_i,\mathbf x_j) < d(\mathbf x_i,ne(\mathbf x_i))\},
\end{equation}
where $ne(\mathbf x_i)$ is the nearest enemy from example $\mathbf x_i$. 
Figure \ref{fig:LSA} illustrates the local set of a particular example ($\mathbf x$, in gray) in a dataset.

The cardinality of the LS of an example indicates its proximity to the decision boundary and also the narrowness of the gap between the classes. Therefore, the LS cardinality will be lower for examples separated from the other class with a narrow margin.  According to \citet{leyvaset}, a high number of low-cardinality local sets in a dataset suggests that the space between classes is narrow and irregular, that is, the boundary is more complex. The local set average cardinality measure (LSC) is calculated here as:
\begin{equation}\label{eq:LSC}
LSC = 1 - \frac{1}{n^2}\sum_{i=1}^{n}{|LS(\mathbf x_i)|},
\end{equation}
where $|LS(\mathbf x_i)|$ is the cardinality of the local set for example $\mathbf x_i$. This measure can complement N1 and L1 by also revealing the narrowness of the between-class margin. Higher values are expected for more complex datasets, in which each example is nearest to an enemy than to other examples from the same class. In that case, each example will have a local set of cardinality 1, resulting in a LSC of $1-\frac{1}{n}$. The asymptotic cost of LSC is dominated by the computation of pairwise distances between all examples, resulting in $O(m \cdot n^2)$ operations.

\subsection{Network Measures}

\citet{Morais2013} and \citet{GarciaEtAl2015} model the dataset as a graph and extract measures for the statistical characterization of complex networks \cite{Kolaczyk2009} from this representation. 
In \citet{GarciaEtAl2015} low correlation values were observed between the basic complexity measures of \citet{HoBasu2002} and the graph-based measures, which supports the relevance of exploring this alternative representation of the data structure. In this paper, we highlight the best measures for the data complexity induced by label noise imputation (\citet{GarciaEtAl2015}), with an emphasis on those with low correlation between each other. 

\begin{figure}[ht!]
   \centering
   \subfloat[Building the graph (unsupervised)]{
        \label{fig:preEnn}
        \includegraphics[scale=0.3]{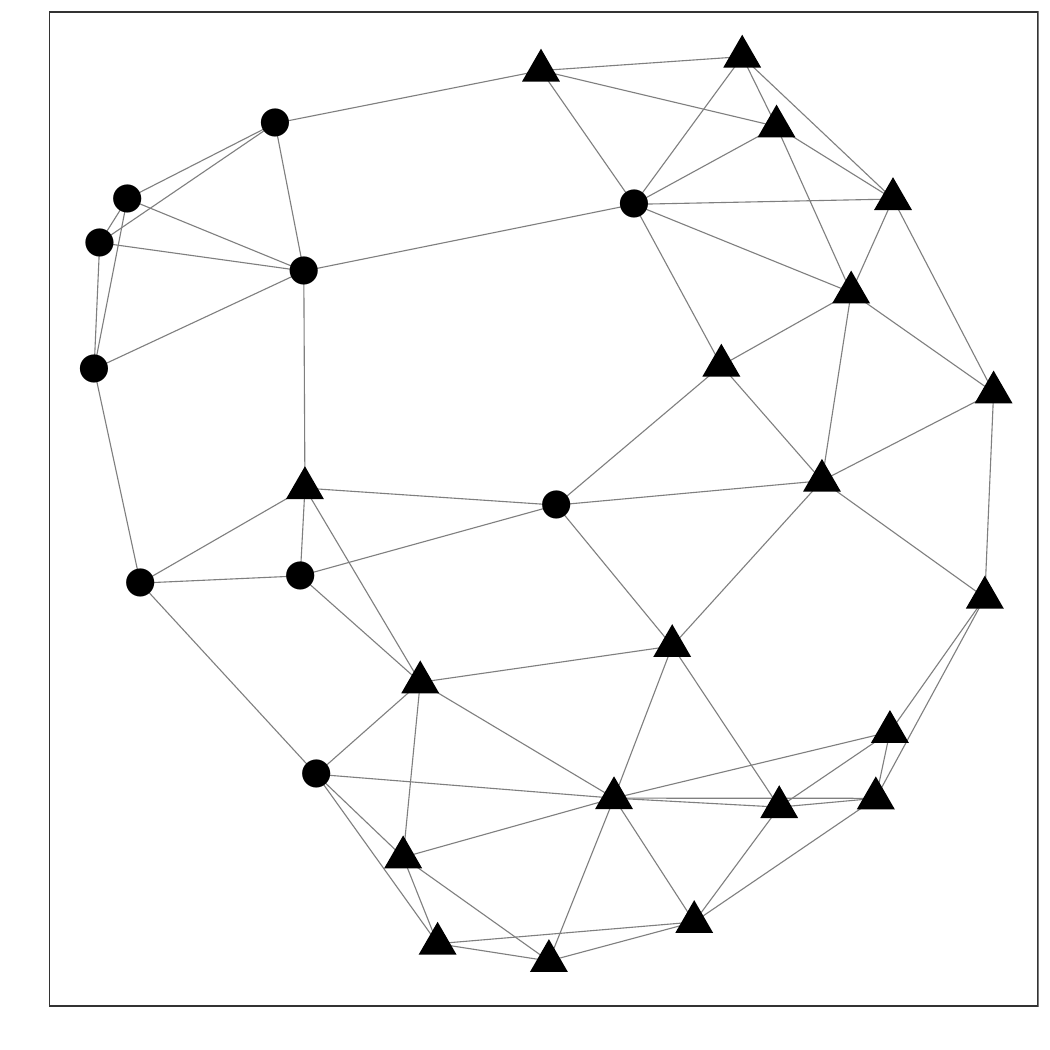}}			
   \subfloat[Pruning process (supervised)]{
        \label{fig:posEnn}
        \includegraphics[scale=0.3]{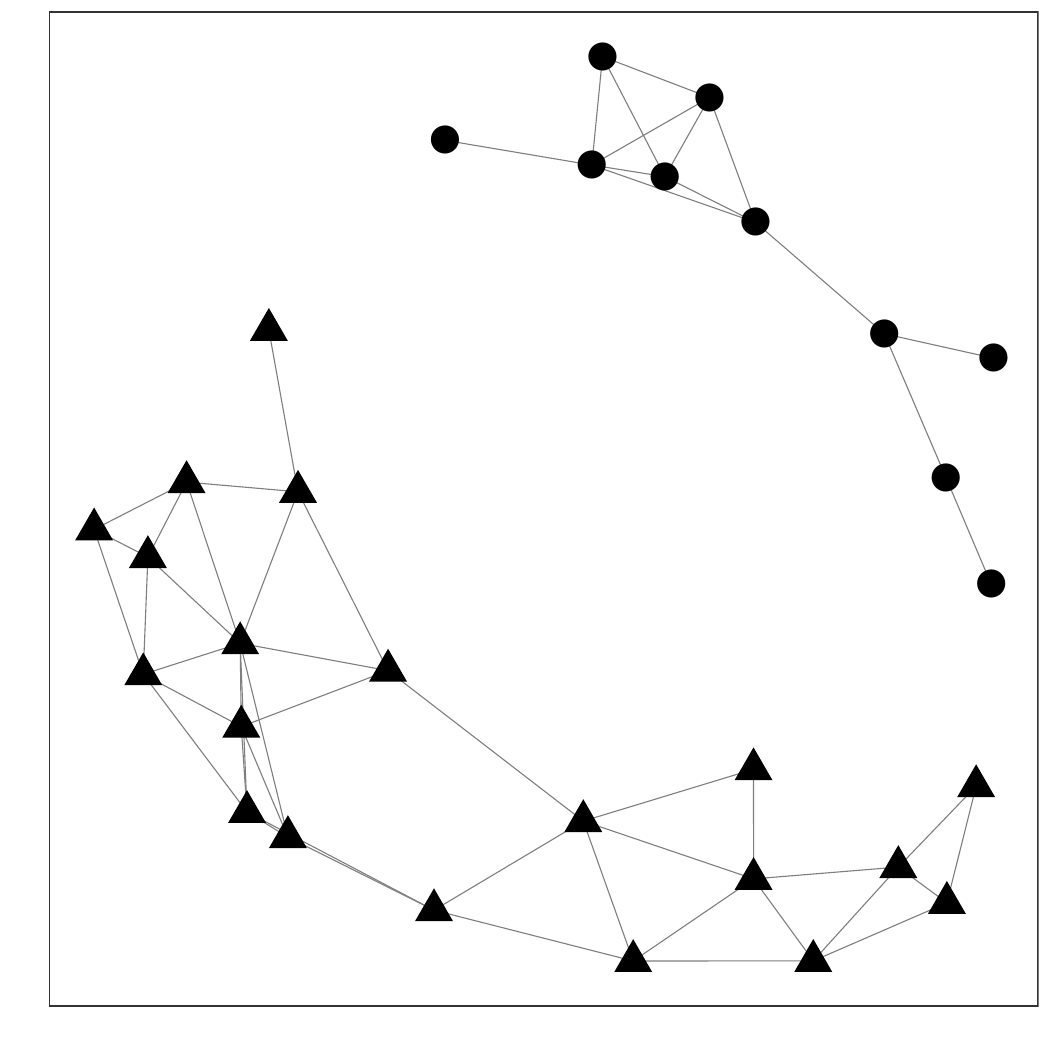}}			
   \caption{Building a graph using $\epsilon$-NN.}
   \label{fig:graph}
\end{figure}

To use these measures, it is necessary to represent the classification dataset as a graph. The obtained graph must preserve the similarities or distances between examples for modeling the data relationships. Each example from the dataset corresponds to a node or vertex of the graph, whilst undirected edges connect pairs of examples and are weighted by the distances between the examples. As in the neighborhood measures, the Gower distance is employed. Two nodes $i$ and $j$ are connected only if $dist(i,j) < \epsilon$. This corresponds to the $\epsilon$-NN method for building a graph from a dataset in the attribute-value format \citep{Zhu2005}.  As in \citet{Morais2013} and \citet{GarciaEtAl2015}, in ECoL the $\epsilon$ value is set to 0.15 (note that the Gower distance is normalized to the range [0,1]).  Next, a post-processing step is applied to the graph, pruning edges between examples of different classes. Figure \ref{fig:graph} illustrates the graph building process for the dataset from Figure \ref{fig:F2}. Figure \ref{fig:preEnn} shows the first step, when the pairs of vertices with $dist(\mathbf x_i,\mathbf x_j) < \epsilon$ are connected. This first step is unsupervised, since it disregards the labels of connected points. Figure \ref{fig:posEnn} shows the graph obtained after the pruning process is applied to disconnect examples from different classes. This step can be regarded as supervised, in which the label information is taken into account to obtain the final graph.

For a given dataset, let $G=(V,E)$ denote the graph built by this process. By construction, $|V| = n$ and $0 \leq |E| \leq \frac{n(n-1)}{2}$. 
Let the $i$-th vertex of the graph be denoted as $v_i$ and an edge between two vertices $v_i$ and $v_j$ be denoted as $e_{ij}$. The extracted measures are described next.  All the measures from this category require building a graph based on the distance matrix between all pairs of elements, which requires $O(m \cdot n^2)$ operations. 
The asymptotic cost of all the presented measures is dominated by the computation of this matrix.

\subsubsection{Average density of the network (Density)}

This measure considers
the number of edges that are retained in the graph built from the dataset normalized by the maximum number of edges between $n$ pairs of data points.
\begin{equation}
Density = 1- \frac{2|E|}{n(n-1)}
\end{equation} 

Lower values for this measure are obtained for dense graphs, in which many examples get connected. This will be the case for datasets with dense regions from a same class in the dataset. 
This type of dataset can be regarded as having lower complexity. On the other hand, a low number of edges will be observed for datasets of low density (examples are far apart in the input space) and/or for which examples of opposite classes are near each other,  implying a higher classification complexity. 

\subsubsection{Clustering coefficient (ClsCoef)} 

The clustering coefficient
of a vertex $v_i$ is given by the ratio of the number of edges between its neighbors and the maximum number of edges that could possibly exist between them. We take as a complexity measure the value:
\begin{equation}\label{eq:cls}
ClsCoef = 1-  \frac{1}{n}\sum_{i=1}^{n}{\frac{2|e_{jk}:v_j,v_k \in N_i|}{k_i(k_i-1)}},
\end{equation}
where 
$N_i = \{v_j: e_{ij} \in E\}$ denotes the neighborhood set of a vertex $v_i$ (those nodes directly connected to $v_i$) and $k_i$ is the size of $N_i$. The sum calculates, for each vertex $v_i$, the ratio of existent edges between its neighbors by the total number of edges that could possibly be formed.  

The Clustering coefficient measure assesses the grouping tendency of the graph vertexes, by monitoring how close to form cliques neighborhood vertexes are. As computed by Equation \ref{eq:cls}, it will be smaller for simpler datasets, which will tend to have dense connections among examples from the same class.

\subsubsection{Hub score (Hubs)} 

The hub score
scores each node by the number of connections it has to other nodes, weighted by the number of connections these neighbors have. Herewith, highly connected vertexes which are also connected to highly connected vertices will have a larger hub score. This is a measure of the influence of each node of the graph. Here we take the formulation:
\begin{equation}\label{eq:hubs}
Hubs = 1- \frac{1}{n}\sum_{i=1}^{n}hub(v_i)
\end{equation}
The values of $hub(v_i)$ are given by the principal eigenvector of $A^tA$, where $A$ is the adjacency matrix of the graph. Here, we take an average value for all vertices.

In complex datasets, in which a high overlapping of the classes is observed, strong vertexes will tend to be less connected to strong neighbors. On the other hand, for simple datasets there will be dense regions within the classes and higher hub scores. Therefore, according to Equation \ref{eq:hubs} smaller Hubs values are expected for simpler datasets.

\subsection{Dimensionality Measures}

The measures from this category give an indicative of data sparsity. They are based on the dimensionality of the datasets, either original or reduced. The idea is that it can be more difficult to extract good models from sparse datasets, due to the probable presence of regions of low density that will be arbitrarily classified.

\subsubsection{Average number of features per points (T2)}

Originally, T2 divides the number of examples in the dataset by their dimensionality  \citep{BasuHo2006}. In this paper we take the inverse of this formulation in order to obtain higher values for more complex datasets, so that:
\begin{equation}
\label{eq:T2}
T2 = \frac{m}{n}
\end{equation}

T2 can be computed at $O(m+n)$. In some work the logarithmic function is applied to the measure (ex. \citet{LorenaEtAl2012}) because T2 can take arbitrarily large or small values. Though, this can take the measure into negative values when the number of 
examples is larger than the number of features.

T2 reflects the data sparsity. If there are many predictive attributes and few data points, they will be probably sparsely distributed in the input space. The presence of low density regions will hinder the induction of an adequate classification model. Therefore, lower T2 values indicate less sparsity and therefore simpler problems.

\subsubsection{Average number of 
PCA dimensions per points (T3)}

The 
measure T3 \citep{LorenaEtAl2012} is defined with a Principal Component Analysis (PCA) of the dataset. Instead of the raw dimensionality of the feature vector (as in T2), T3 uses the number of PCA components needed to represent 95\% of data variability ($m'$) as the base of data sparsity assessment. The measure is calculated as:
\begin{equation}
T3 = \frac{m'}{n}
\end{equation}

The value $m'$ can be regarded as an estimate of the intrinsic dataset dimensionality after the correlation among features is minimized.   
As in the case of T2, smaller values will be obtained for simpler datasets, which will be less sparse. Since this measure requires performing a PCA analysis of the dataset, its worst cost is $O(m^2 \cdot n + m^3)$.

\subsubsection{Ratio of the PCA Dimension to the Original Dimension (T4)}

This measure 
gives a rough measure of the proportion of relevant dimensions for the dataset \citep{LorenaEtAl2012}. This relevance is measured according to the PCA criterion, which seeks a transformation of the features to uncorrelated linear functions of them that are able to describe most of the data variability. T4 can be expressed by:
\begin{equation}
T4 = \frac{m'}{m}
\end{equation}
The larger the T4 value, the more of the original features are needed to describe data variability. This indicates a more complex relationship of the input variables. The asymptotic cost of the measure is $O(m^2 \cdot n + m^3)$.

\subsection{Class Imbalance Measures}

These measures try to capture one aspect that may largely influence the predictive performance of ML techniques when solving data classification problems: class imbalance, that is, a large difference in the number of examples per class in the training dataset. Indeed, when the differences are severe, most of the ML classification techniques tend to favor the majority class and present generalization problems. 

In this section we present some measures for capturing class imbalance. If the problem has a high imbalance in the proportion of examples per class, it can be considered more complex than a problem for which the proportions are similar. 

\subsubsection{Entropy of class proportions (C1)} 

The C1 measure was used in \citet{LorenaEtAl2012} to capture the imbalance in a dataset. It can be expressed as:
\begin{equation}
C1 = 1 + \frac{1}{\log(n_{c_i})}\sum_{i=1}^{n_c} p_{c_i} \log(p_{c_i}),
\end{equation}
where $p_{c_i}$ 
is the proportion of examples in each of the classes. This measure will achieve minimum value for balanced problems, that is, problems in which all proportions are equal. These can be considered simpler problems according to the class balance aspect. The asymptotic cost for computing this measure is $O(n)$ for obtaining the proportions of examples per class.

\subsubsection{Imbalance ratio (C2)} 

The 
C2 measure is a well known index computed for measuring class balance. Here we adopt a version of the measure that is also suited for multiclass classification problems \citep{tanwani2010classification}:
\begin{equation}
C2 = 1 - \frac{1}{IR},
\end{equation}
where:
\begin{equation}
IR = \frac{n_c-1}{n_c}\sum_{i=1}^{n_c}{\frac{n_{c_i}}{n-n_{c_i}}},
\end{equation}
where $n_{c_i}$ is the number of instances from the $i$-th class. These numbers can be computed at $O(n)$ operations.
Larger values of C2 are obtained for imbalanced problems. The minimum value of C2 
is achieved for balanced problems, in which $n_i=n_j$ for all $i,j = 1, \ldots, n_c$.

%% file: Others.tex
\subsection{Other Measures}

This section gives an overview of some other measures that can be used to characterize the complexity of classification problems found in the relate literature. Part of these 
measures 
was not formally included previously because they capture similar aspects already measured by the described measures. Other measures were excluded because they have a high computational cost.

\citet{van2007measures} 
present some variations of the T1 measure. One of them is quite similar to the LSC measure, with a difference on the normalization used by LSC. 
Another variation first generates an MST connecting the hyperspheres centers given by T1 and 
then counts the number of vertexes that connect examples from different classes. 
There is also a measure that computes the density of the hyperspheres. We believe that the LSC measure complements T1 at a lower computational cost.

\citet{sotoca2006meta} present 
some density measures. The first one, named D1, gives the {\em average number of examples per unit of volume} in the dataset. 
The {\em volume of local neighborhood} (D2) measure gives the average volume occupied by the $k$ nearest neighbors of each example. Finally, the {\em class density in overlap region} (D3) determines the density of each class in the overlap regions. It counts, for each class, the number of points lying in the same region of a different class. 
Although these measures give an overview of data density, we believe that they do not allow to extract complementary views of the problem complexity already captured by the original neighborhood-based measures. Furthermore, they may have a higher computational cost and present an additional parameter (e.g. the $k$ in $k$ nearest neighbors) to be tuned.

Some of the measures found in the literature propose to analyze the dataset using a divisive approach or in multiple resolutions. Usually they show a high computational cost, that can be prohibitive for datasets with a moderate number of features. \citet{Singh2003} reports some 
of such measures. 
Their partitioning algorithm generates hypercuboids in the space, at different resolutions (with increasing numbers of intervals per feature from 0 to 31). At each resolution, the data points are assigned into cells. 
{\em Purity} measures whether the cells contain examples from a same class or from mixed classes. 
The {\em nearest neighbor separability} measure counts, for each example of a cell, the proportion of its nearest neighbors that share its class. The cell  measurements are linearly weighted to obtain a single estimate and  
 the overall measurement across all cells at a given resolution is exponentially weighted. 
Afterwards, 
the area under the curve defined by one separability measure versus the resolution defines the overall data separability. 
In \citet{Singh2003PRISM} two more measures based on the space partitioning algorithm 
are defined: {\em collective entropy}, which is the level of uncertainty accumulated at different resolutions; and \textit{data compactness}, related to the proportion of non-empty cells at different resolutions. 

In \citet{armano2016experimenting} a method named {\em Multi-resolution Complexity Analysis} (MRCA) is used to partition a dataset. 
Like in T1, hyperspheres of different amplitudes are drawn around the examples and the imbalance regarding how many examples of different classes they contain is measured. A new dataset of profile patterns is obtained, which is clustered. 
Afterwards, each cluster is evaluated and ranked according to a complexity metric called {\em Multiresolution Index} (MRI). 

\citet{armano2015direct} presents how to obtain a class signature which can be used to identify, for instance, the discriminative capability of the input features. This could be regarded as a feature-based complexity measure, although more developments are necessary, since the initial studies considered binary-valued features only.

\citet{mthembu2008note} present a {\em Separability Index} SI, which takes into account the average
number of examples in a dataset that have a nearest neighbor with the same label. This is quite similar to what is captured by N3, except for using more neighbors in NN classification. Another measure named {\em Hypothesis margin} (HM) takes the distance between the nearest neighbor of an object of the same class and a nearest enemy of another class. This largely resembles the N2 computation. 

Similarly to D3, \citet{sotoca2006meta} and \citet{anwar2014measurement} introduce a complexity measure which also focuses on local information for each example by employing the nearest neighbor algorithm. 
If the majority of the $k$ nearest neighbors of an example share its label, this point can be regarded as easy to classify. Otherwise, it is a difficult point. An overall complexity measure is given by the proportion of data points classified as difficult. 

\citet{leyvaset} define some measures based on the concept of Local Sets previously described, which employ neighborhood information. 
Besides LSC, \citet{leyvaset} also propose to cluster the data in the local sets and then count the number of obtained clusters. This measure is related to T1. The third measure is named {\em number of invasive points} (Ipoints), which uses the local sets to identify borderline instances and is related to N1, N2 and N3. 

\citet{smith2014instance} propose a 
set of measures devoted to understand why some data points are harder to classify than others. They are called \textit{instance hardness} measures. One advantage of such approach is to reveal the difficulty of a problem at the instance level, rather than at the aggregate level with the entire dataset. Nonetheless, the measures can be averaged to give an estimate at the dataset level. As shown in some recent work on dynamic classifier selection \citep{cruz2017dynamic,cruz2018dynamic}, the concept of instance hardness is very useful for pointing out classifiers able to perform well in confusing or overlapping areas of the dataset, giving indicatives of a local level of competence. Most of the complexity measures previously presented, although formulated for obtaining a complexity estimate per dataset, can be adapted in order to assess the contribution of each example to the overall problem difficulty. Nonetheless, this is beyond the scope of this paper.

One very effective instance hardness measure from \citep{smith2014instance} is the 
{\em $k$-Disagreeing Neighbors} ($k$DN), which gives the percentage of the $k$ nearest neighbors that do not share the label of an example. This same concept was already explored in the works of  \citet{SotocaEtAl2005,mthembu2008note,anwar2014measurement}. The  
{\em Disjunct Size} (DS) corresponds to the size of a {\em disjunct} that covers an example divided by the largest disjunct produced, in which disjuncts are obtained using the C4.5 learning algorithm.
A relate measure is the {\em Disjunct Class Percentage} (DCP), which is the number of data points in a disjunct that belong to a same class divided by the total number of examples in the disjunct. 
The {\em Tree Depth} (TD) returns the depth of the leaf node that classifies an instance in a decision tree. 
The previous measures give estimates from the perspective of a decision tree classifier. 
In addition, the {\em Minority Value} (MV) index is the ratio of examples sharing the same label of an example to the number of examples in the majority class.  
The {\em Class Balance} (CB) index presents an alternative to measuring the class skew. The C1 and C2 measures previously described are simple alternatives already able to capture the class imbalance aspect. 

\citet{elizondo2012linear}
focus their study on the relationship between linear separability and the level of complexity of classification datasets. Their method 
uses {\em Recursive Deterministic Perceptron} (RDP) models and counts the 
number of hyperplanes needed to transform the original problem, which may not be linearly separable, into a linearly separable problem. 

In \citet{skrypnyk2011irrelevant} various class separability measures are presented, focusing on 
feature selection. Some parametric measures are the \textit{Mahalanobis} and the \textit{Bhattacharyya} distances between the classes and the {\em Normal Information Radius}. 
These measures are computationally intensive due to the need to compute covariance matrices and their inverse. An information theoretic measure is the \textit{Kullback-Leibler} distance. It quantifies the discrepancy between two probability distributions. Based on discriminant analysis, a number of class separability measures can also be defined. 
This family of techniques is closely related to measures F1v and N2 discussed in this survey. 

\citet{Cummins2013} 
also defines some alternative complexity measures. The first, named N5, consists of multiplying N1 by N2. According to \citet{fornells2007methodology}, the multiplication of N1 and N2 emphasizes extreme behavior concerning class separability. 
Another measure (named {\em Case Base Complexity Profile}) retrieves the $k$ nearest neighbors of an example $\mathbf x$ for increasing values of $k$, from 1 up to a limit $K$. At each round, the proportion of neighbors that have the same label as $\mathbf x$ is counted. The obtained values are then averaged. 
Although interesting, this measure can be considered quite costly to compute. 

More recently, \citet{zubek2016complexity} presented a complexity curve based on the \textit{Hellinger} distance of probability distributions, assuming that the input features 
are independent. It takes subsets of different sizes from a dataset and verifies if their information content is similar to that of the original dataset. The computed values are plotted and the area under the obtained curve is used as an estimate of data complexity. The proposed measure is also applied in data pruning. 
The measure values computed turned out to be quite correlated to T2.

In the recent literature there are also studies on generalizations of the complexity measures for other types of problems. In \citet{lorena2018meta} these measures are adapted to quantify the difficulty of regression problems. \citet{charte2016impact}  present a complexity score for multi-label classification problems. \citet{smith2009cross}  surveys some strategies for measuring the difficulty of optimization problems.

%% file: ECol.tex
\section{The ECol Package}\label{cap:core}

Based on the review performed, we assembled a set of 22 complexity measures into an R package named ECoL (\textit{Extended Complexity Library}), available at
 CRAN\footnote{\url{https://cran.r-project.org/package=ECoL}} and GitHub\footnote{\url{https://github.com/lpfgarcia/ECoL}}. Table \ref{tab:medidas} summarizes the characteristics of the complexity measures included in the package. 
It presents the category, name, acronym, 
and the limit values (minimum and maximum) assumed by these measures. Taking the measure F1 as an example, according to Table \ref{tab:medidas} 
its higher limit is 1 (when the average 
values of the attributes are the same for all classes), its lower value is approximately null. 
In our implementations, all measures assume values that are in bounded intervals. Moreover, for all measures, the higher the value, the greater the complexity measured. 
We also present the worst case asymptotic time complexity cost for computing the measures, where $n$ stands for the number of points in a dataset, $m$ corresponds to its number of features, $n_c$ is the number of classes and $l$ is the number of novel points generated in the case of the measures L3 and N4. All distance- and network-based measures are based on information from a distance matrix between all pairs of examples of the dataset, which can  be computed only once and reused for obtaining the values of all those measures. The same reasoning applies to the linearity measures, since all of them involve training a linear SVM, from which the required information for computing the individual measures can be obtained.

\begin{table}[htb]
\caption{Characteristics of the complexity measures.}
\centering
\label{tab:medidas}
\scalebox{0.80}{
\begin{tabular}{l|l|l|c|c|c}
\hline
\textbf{Category} & \textbf{Name} & \textbf{Acronym} & \textbf{Min} & \textbf{Max} 
& \textbf{Asymptotic cost}\\
\hline \hline
\multirow{5}{*}{Feature-based} 
& Maximum Fisher's discriminant ratio& F1            &  $\approx 0$           & 1  
& 
$O(m\cdot n)$   \\ \cline{2-6}
& Directional vector maximum Fisher's discriminant ratio& F1v            &  $\approx 0$           & 1 
& $O(m \cdot n \cdot n_c + m^3 \cdot n_c^2)$   \\ \cline{2-6}
& Volume of overlapping region &F2            & $0$           & $1$     
& $O(m \cdot n \cdot n_c)$\\ \cline{2-6}
& Maximum individual feature efficiency& F3            & $0$           & $1$
& 
$O(m \cdot n \cdot n_c)$ \\ \cline{2-6}
& Collective feature efficiency& F4          &$0$  & $1$           & $O(m^2 \cdot n \cdot n_c)$\\ \cline{2-6}
\hline
\multirow{3}{*}{Linearity} & Sum of the error distance by linear programming
& L1            & $0$           & $\approx 1$    
& $O(n^2)$\\ \cline{2-6}
& Error rate of linear classifier & L2            & $0$           & $1$         
& 
$O(n^2)$\\ \cline{2-6}
&Non linearity of linear classifier& L3            & $0$           & $1$           & 
$O(n^2 + m \cdot l \cdot n_c)$ \\ \hline
\multirow{6}{*}{Neighborhood}
& Faction of borderline points & N1      &       $0$           & $1$           & $O(m \cdot n^2)$\\ \cline{2-6}
& Ratio of intra/extra class NN distance & N2            & $0$           & $\approx 1$
&$O(m \cdot n^2)$ \\ \cline{2-6}
&Error rate of NN classifier & N3            & $0$           & $1$  
& $O(m \cdot n^2)$\\ 
\cline{2-6}
& Non linearity of NN classifier& N4            & $0$           & $1$           
& $O(m \cdot n^2 + m \cdot l \cdot n)$\\ \cline{2-6}
& Fraction of hyperspheres covering data  & T1            & $0$           & $1$ 
&$O(m \cdot n^2)$ \\ \cline{2-6}
& Local set average cardinality & LSC & 0 & $1 - \frac{1}{n}$
& $O(m \cdot n^2)$\\
\hline
\multirow{3}{*}{Network}
& Density       & Density & $0$           & $1$           & 
$O(m \cdot n^2)$ \\ \cline{2-6}
& Clustering Coefficient & ClsCoef       & $0$           & $1$           
& $O(m \cdot n^2)$ \\ \cline{2-6}
& Hubs          & Hubs &$0$           & $1$           
& $O(m \cdot n^2)$ \\ \cline{2-6}
\hline
\multirow{3}{*}{Dimensionality}
&Average number of features per points & T2            & $\approx 0$           & $m$           
& $O(m+n)$\\ \cline{2-6}
&Average number of PCA dimensions per points & T3            & $\approx 0$           & $m$        
& $O(m^2 \cdot n + m^3)$\\ \cline{2-6}
&Ratio of the PCA dimension to the original dimension & T4            & 0 & 1   
& $O(m^2 \cdot n + m^3)$\\ \cline{2-6}
\hline
\multirow{2}{*}{Class imbalance} &
Entropy of classes proportions & C1 & 0 & 1 
& $O(n)$\\ \cline{2-6}
&
Imbalance ratio & C2 & 0 & 1 &
$O(n)$\\ \hline
\end{tabular}
}
\end{table}

Another relevant observation is that although each measure gives an indication into the complexity of the problem according to some characteristics of its learning dataset, a unified interpretation of their values is not easy. Each measurement has an associated limitation (for example, the feature separability measures cannot cope with situations where an attribute has different ranges of values for the same class - Figure \ref{fig:probF2c}) and must then be considered only as an estimate of the problem complexity, which may have associated errors. Since the measures are made on a dataset $T$, they also give only an apparent measurement of the problem complexity \citep{HoBasu2002}. This reinforces the need to analyze the measures together to provide more robustness to the reached conclusions. There are also cases where some caution must be taken, such as in the case of F2, whose final values depend on the number of predictive attributes in the dataset. 
This particular issue is pointed out by \citet{Singh2003PRISM}, which states that the complexity measures  ideally should be conceptually uncorrelated to the number of features, classes, or number of data points a dataset has, making the complexity measure values for different datasets more comparable. 
This requirement is clearly not fulfilled by 
F2. Nonetheless, for the dimensionality- and balance-based measures, \citet{Singh2003}'s assertion does not apply, since they are indeed concerned with the relationship of the numbers of dimensions and data points a dataset has.  

For instance, a linearly separable problem with an oblique hyperplane will have high F1, indicating that it is complex, and also a low L1, denoting that it is simple. LSC, on the other hand, will assume a low value for a very imbalanced two-class dataset in which one of the classes contains one unique example and the other class is far and densely distributed. This would be an indicative of a simple classification problem according to LSC interpretation, but data imbalance should be considered too. In the particular case of class imbalance measures, \citet{batista2004study} show that the harmful effects due to class imbalance are more pronounced when there is also a large overlap between the classes. Therefore, 
these measures should be analyzed together with measures able to capture class overlap (ex. C2 with N1).  Regarding network-based measures, the $\epsilon$ parameter in the $\epsilon-NN$ algorithm in ECol is fixed at 0.15, although we can expect that different values may be more appropriate for distinct datasets. With the free distribution of the ECol package, interested users are able to modify this value and also other parameters (such as the distance metric employed in various measures) and test their influence in the reported results.

Finally, whist some measures are based on classification models derived from the data, others use only statistics directly derived from the data. Those that use classification models, i.e., a linear classifier or an NN classifier, are: F1v, L1, L2, L3,  N3, N4. This makes these measures dependent on the classifier decisions they are based on, which in turn depends on some choices in building the classifiers, such as the algorithm to derive a linear classifier, or the distance used in nearest-neighbor classification \citep{bernado2005domain}. 
Other measures are based on characteristics extracted from the data only, although the N1 and the network indexes involve pre-computing a distance-based graph from the dataset. Moreover, it should be noticed that all measures requiring the computation of covariances or (pseudo-)inverses are time consuming, such as F1v, T3 and T4.

\citet{smith2014instance} highlight another notice-worthy issue that some measures are unable to provide an instance-level hardness estimate. Understanding which instances are hard to classify may be valuable information, since more efforts can be devoted to them. However, many of the complexity measures originally proposed for a dataset-level analysis can be easily adapted to give instance-level hardness estimates. This is the case of N2, which averages the intra- and inter-class distances from each example to their nearest neighbors. 

%% file: Aplications.tex
\section{Application Areas}\label{cap:apli}

The data complexity measures have been applied to support of various supervised ML tasks. This section discusses some of the main applications of the complexity measures found in the relate literature. They can be roughly divided into the following categories: 

\begin{enumerate}
\item data analysis, where the measures are used to understand the peculiarities of a particular dataset or domain;

\item data pre-processing, where the measures are employed to guide data-preprocessing tasks; 

\item learning algorithms, where the measures are employed for understanding or in the design of ML algorithms;

\item meta-learning, where the measures are used in the meta-analysis of classification problems, such as in choosing a particular classifier.

\end{enumerate}

\subsection{Data Analysis}

Following the data analysis framework, some works employ the measures to better understand how the main characteristics of datasets available for learning in an application domain affect the achievable classification performance. 
For instance, in \citet{LorenaEtAl2012} the complexity measures are employed to analyze the characteristics of cancer gene expression data that have most impact on the predictive performance in their classification. The measures which turned out to be the most effective in such characterization were: T2 and T3 (data sparsity), C1 (class imbalance), F1 (feature-based) and N1, N2 and N3 (neighborhood-based). The complexity measures values were also monitored after a simple feature selection strategy, which revealed the importance of such pre-processing in reducing the complexity of those high-dimensional classification problems. More recently, \citet{moran2017can} conducted a similar study with more classification and feature selection techniques and reached similar conclusions. They also tried to answer whether classification performance could be predicted by the complexity measure values in the case of the microarray datasets. In that study, the complexity measures with highlighted results were: F1 and F3 (feature-based), N1 and N2 (neighborhood-based) when the k-nearest neighbor (kNN) classifier is used and L1 (linearity-based) in the case of linear classifiers.

Another interesting use of the data complexity measures has been in generating artificial datasets with controlled characteristics. This resulted in some data repositories with systematic coverage for evaluating classifiers under different challenging conditions  \citep{macia2010search,smith2014easy,macia2014towards,de2018using}. 
In \citep{macia2010search} a multi-objective Genetic Algorithm (GA) is employed to select subsets of instances of a dataset targetting at specific ranges of values of one or more complexity measures. In their experiments, one representative of each of the categories of \citet{HoBasu2002}'s complexity measures was chosen to be optimized: F2, N4 and T1. Later, in 
\citep{macia2014towards} the same authors analyze the UCI repository. 
They experimentally observed that the majority of the UCI problems are easy to learn (only 3\% were challenging for the classifiers tested). To increase the diversity of the repository, \citet{macia2014towards} suggest to include artificial datasets carefully designed to span the complexity space, which are produced by their multiobjective GA. This gave rise to the UCI+ repository. In \citep{de2018using}, a hill-climbing algorithm is also employed to find synthetic datasets with targetted complexity measure values. Some measures devoted to evaluate the overlapping of the classes were chosen to be optimized: F1, N1 and N3. The algorithm starts with randomly produced datasets and the labels of the examples are iteratively switched seeking to reach a given complexity measure value.

\subsection{Data Pre-Proprocessing}

The data complexity measures have also been used to guide data pre-processing tasks, such as Feature Selection (FS) \citep{LiuEtAl2010}, noise identification \citep{Frenay2013} and dealing with data imbalance \citep{he2009learning,fernandez2018learning}.

In FS, the measures have been used to both guide the search for the best featues in a dataset \citep{Singh2003PRISM,okimoto2017} or to 
understand feature selection effects \citep{baumgartner2006complexity,PranckevicieneEtAl2006,skrypnyk2011irrelevant}. For instance, \citet{PranckevicieneEtAl2006} propose to quantify whether FS effectively changes the complexity of the original classification problem. 
They found that FS was able to increase class separability in the reduced spaces, as measured by N1, N2 and T1. \citet{okimoto2017} assert the power of some complexity measures in ranking the features contained in synthetic datasets for which the relevant features are known a priori. As expected, feature-based measures (mainly F1) are very effective in revealing the relevant features in a dataset, although some neighborhood measures (N1 and N2) also present highlighted results. Another interesting recent work on feature selection uses a combination of the feature-based complexity measures F1, F2 and F3 to support the choice of thresholds in the number of features to be selected by FS algorithms \citep{seijo2019developing}.

Instance (or prototype) selection (IS) has also been the theme of various works involving the data complexity measures. In one of the first works in  the area, \citet{MollinedaEtAl2005} tries to predict which instance selection algorithm should be applied to a new dataset. They report highlighted results of the F1 measure in identifying situations in which an IS technique is needed. Other works include: \citet{leyvaset} and \citet{cummins2011dataset}. \citet{leyvaset}, for instance, presents some complexity measures which are claimed to be  specifically designed for characterizing IS problems. Among them is the LCS measure.  
\citet{kim2009using} perform a different analysis. They are interested in investigating whether the complexity measures can be calculated at reduced datasets while still preserving the characteristics found in the original datasets. Only separability-measures are considered, among them F1, F2, F3 and N2. The results were positive for all measures, except for F1.

Under his partitioning framework, \citet{Singh2003PRISM} 
discusses how potential outliers can be identified in a dataset. 
Other uses of the complexity measures in the noise identification context are: \citep{smith2014instance,SaezEtAl2013,garcia2013noisy,GarciaEtAl2015,GarciaEtAl2015b}. 
\citet{GarciaEtAl2015}, for example, investigate how different label noise levels affect the values of the complexity measures. Neighborhood-based (N1, N2 and N3), feature-based (F1 and F3) and some network-based measures (density and hubs) were found to be effective in capturing the presence of label noise in classification datasets.  
Two of the measures most sensitive to noise imputation were then combined to develop a new noise filter, named {\em GraphNN}. 

\citet{gong2012kolmogorov} 
found that the data complexity of a classification problem is more determinant in model performance than class imbalance, and that class imbalance amplifies the effects of data complexity.  
\citet{vorraboot2012modified} adapted the back-propagation (BP) algorithm to take into account the class overlap and the imbalance ratio of a dataset, using the F1 feature-based measure and the imbalance-ratio for binary problems. 
\citet{lopez2012analysis} 
uses the F1 measure to analyze the 
differences between pre-processing techniques and cost-sensitive learning for addressing  imbalanced data classification. 
Other works in the analysis of imbalanced classification problems include \citet{xing2013preliminary}, \citet{anwar2014measurement}, \citet{santos2018cross} and \citet{ZHANG2019204}. More discussions on the effectiveness of data complexity analysis related to the data imbalance theme can be found in \citep{fernandez2018learning}.

\subsection{Learning Algorithms}

Data complexity measures can also be employed for analysis at the level of algorithms. These analyses can be for devising, tuning or understanding the behavior of different learning algorithms. 
For instance, \citet{zhao2016initial} use the complexity measures to understand the data transformations performed by \textit{Extreme Learning Machines} at each of their layers. They have noticed some small changes in the complexity as measured by F1, F3 and N2, which were regarded as non-significant.

A very popular use of the data complexity measures is to outline the domains of competence of one or more ML algorithms \citep{luengo2015automatic}. This type of analysis allows to identify problem characteristics for which a given technique will probably succeed or fail. While improving the understanding of the capabilities and limitations of each technique, it also supports the choice of a particular technique for solving a new problem. 
It is possible to reformulate a learning procedure by taking into account the complexity measures too, or to devise new ML and pre-processing techniques. 

In the analysis of the domains of competence of algorithms, one can cite: \citet{ho2000complexity,Ho2002} for \textit{random decision forests}; 
\citet{bernado2005domain} for the 
\textit{XCS} classifier;  
\citet{HoMansilla2006} for \textit{NN, Linear Classifier, Decision Tree, Subspace Decision Forest} and \textit{Subsample Decision Forest}; 
\citet{flores2014domains} 
for finding datasets that fit for a \textit{semi-naive Bayesian Network Classifier} (BNC) and to 
recommend the best semi-naive BNC to use for a new dataset; \citet{TrujilloEtAl2011} for a \textit{Genetic Programming classifier}; \citet{ciarelli2013impact} for \textit{incremental learning} algorithms; \citet{garcia2012data,fornells2007methodology} for CBR; and \citet{britto2014dynamic} for the \textit{Dynamic Selection} (DS) of classifiers.

In \citet{luengo2015automatic} a general automatic method 
for extracting the domains of competence of any ML classifier is proposed. This is done by monitoring the values of the data complexity measures and relating them to the difference in the training and testing accuracies of the classifiers. 
Rules are extracted from the measures to identify when the classifiers will achieve a good or bad accuracy performance.  

The knowledge advent from the problem complexity analysis can also be used for improving the design of existent ML techniques.
For instance, \citet{smith2014instance} propose a modification of the \textit{back-propagation} algorithm for training Artificial Neural Networks (ANNs) which embed their concept of instance hardness.  
Therein, the error function of the BP algorithm places more emphasis on the hard instances. 
Other works along this line include: \citet{vorraboot2012modified} also on NN, using the measures F1 and imbalance ratio; \citet{campos2012local} in DT ensembles, using the N1 and F4 measures. Recently, \citet{brun2017framework} proposed a framework for dynamic classifier selection in ensembles. It uses a subset of the complexity measures for both: selecting subsets of instances to train the pool of classifiers that compose the ensemble; and to determine the predictions that will be used for a given subproblem, which will favor classifiers trained on subproblems of similar complexity to the query subproblem. They have selected one measure from each of the \citet{HoBasu2002}'s original categories which showed low Pearson correlation with each other to be optimized by a GA suited for DS: F1, N2 and N4.

On the other hand, some works have devised new approaches for data classification based on the information of the complexity measures. This is the case of 
\citet{LorenaCarvalho2010}, in which the measures F1 and F2 are used as splitting criteria for decomposing multiclass problems. \citet{quiterio2018using} also work on the decomposition of multiclass problems, using the complexity measures to place the binary classifiers in \textit{Directed Acyclic Graph} structures. No specific complexity measure among those tested in the paper (namely F1, F3, N1, N2 and T1) could be regarded as best suited for optimizing the DAG structures, although all of them were suitable choices for evaluating the binary classifiers. \citet{sun2019novel} perform hierarchical partitions of the classes minimizing classification complexity, which are estimated according to the measures F1, F2, F3, N2, N3 and a new measure based on centroids introduced in their work. The best experimental results were obtained for the measures F1, F3 and centroid-based.  

Another task that can be supported by the estimates on problem complexity is to tune the parameters of the ML techniques for a given problem.
In \citet{he2015quantification} the 
data complexity measures 
are applied to describe the leak quantification problem. They employ one representative measure of each of the \citet{HoBasu2002}'s categories: F2, N1 and T1. 
In addition, a 
parameter tuning procedure 
which minimizes data complexity under some domain-specific constraints is proposed. Measures N1 and T1 achieved better results. 
\citet{nojima2011meta} use the complexity measures to specify the parameter values of fuzzy classifiers. Some decision rules for binary classification problems based on measures F4, L1, L2, N1 and T2 are reported. N4 is also mentioned as key measure in the case of multiclass problems.

\subsection{Meta-Learning}

In Meta-learning (MtL), meta-knowledge about the solutions of previous problems is used to aid the solution of a new problem \citep{VilaltaDrissi2002}. For this, a meta-dataset composed of datasets for which the solutions are known is usually built. They must be described by meta-features, which is how the complexity measures are mainly used in this area. Some works previously described have made use of meta-learning so they also fall in this category (e.g.,  \citet{smith2014easy,leyvaset,nojima2011meta,ZHANG2019204}).

The work of \citet{sotoca2006meta} is one of the first to 
present a general meta-learning framework based on a number of data complexity measures. \citet{van2007measures} employ the data complexity measures to characterize classification problems in a meta-learning setup designed to predict the expected accuracy of some ML techniques. 
\citet{krijthe2012improving} 
compare classifier selection using cross-validation with meta-learning. 
\citet{cavalcanti2012data} 
use the data complexity measures F1, F2, F3, N1, N2, T1 and T2
to predict the behavior of the NN classifier. In \citet{GarciaEtAl2015b} an MtL recommender system able to predict the expected performance of noise filters in noisy data identification tasks is presented. For such, a meta-learning database is created, containing meta-features, characteristics extracted from several corrupted datasets, along with the performance of some noise filters when applied to these datasets. Along with some standard meta-learning meta-features, the complexity measures N1 and N3 have a higher contribution to the prediction results. More recent works on meta-learning include: \citet{cruz2015meta,das2016meta,roy2016meta,parmezan2017metalearning,cruz2017meta,cruz2018dynamic,garcia2018classifier,shah2018analyzing}. In \citep{garcia2018classifier}, for example, all of the complexity measures described in this work were employed to generate regression models able to  predict the accuracies of four classifiers with very distinct biases: ANN, decision tree, kNN and SVM. The estimated models were effective in such predictions. The top-ranked meta-features chosen by one particular regression technique (Random Forest - RF) were N3, N1, N2, density and T1. All of them regard neighborhood-based information from the data (in the case of density, in the form of a graph built from the data). 

Another interesting usage of the complexity measures in the meta-analysis of classification problems is presented in \citep{munoz2018instance}. There an instance space is built based on meta-features extracted from a large set of classification problems, along with the performance of multiple classifiers. Among the meta-features used are the complexity measures F3, F4, L2, N1 and N4. The instance space framework provides an interesting overview of which are the hardest and easiest datasets and also to identify  strengths and weaknesses of
individual classifiers. The paper also presents a method to generate new datasets which better span the instance space. 

\subsection{Summary}

A summary of the main applications of the data complexity measures found in the literature is presented in Table \ref{tab:applications}. It can be observed that these measures have been mainly employed in the characterization of the domains of competence of various learning and also pre-processing techniques, by revealing when they will perform well or not.  These are generalized to the use of the measures as meta-features for describing datasets in meta-learning studies. 

\begin{table}[htb]
\caption{Some work applying the data complexity measures.}
{\footnotesize
\centering
\label{tab:applications}
\scalebox{0.83}{
\begin{tabular}{l|l|l}
\hline
\textbf{Category} & \textbf{Sub-type} & \textbf{References}  \\
\hline \hline
\multirow{2}{*}{Data Analysis} 
& Domain understanding& \citet{LorenaEtAl2012,kamath2008toward,garcia2016effects}\\
& & \citet{moran2017can}  \\ \cline{2-3}
& \multirow{1}{*}{Data generation}& \citet{macia2014towards,macia2008preliminary,macia2010search,macia2013learner}\\
& & \citep{smith2014easy,de2018using,munoz2018instance} \\
\hline
\multirow{6}{*}{Data Pre-processing} 
& \multirow{2}{*}{Feature Selection} & \citet{Singh2003PRISM,okimoto2017,baumgartner2006complexity,PranckevicieneEtAl2006} \\
& & \citet{skrypnyk2011irrelevant,seijo2019developing} \\
\cline{2-3}
& Instance Selection& \citet{MollinedaEtAl2005,leyvaset,cummins2011dataset,kim2009using}  \\ \cline{2-3}
& Noise identification& \citet{Singh2003PRISM,smith2014instance,SaezEtAl2013,garcia2013noisy,GarciaEtAl2015,GarciaEtAl2015b}  \\ \cline{2-3}
& \multirow{2}{*}{Class imbalance} & \citet{gong2012kolmogorov,vorraboot2012modified,lopez2012analysis,xing2013preliminary}\\
& & \citet{anwar2014measurement,ZHANG2019204,santos2018cross} \\ \hline
\multirow{7}{*}{Learning algorithms} &
 \multirow{3}{*}{Domain of competence}& \citet{bernado2005domain,HoMansilla2006,flores2014domains} \\
 & & \citet{TrujilloEtAl2011,ciarelli2013impact,garcia2012data,fornells2007methodology} \\
 & & \citet{ho2000complexity,britto2014dynamic,luengo2015automatic,lucca2017analyzing}  \\ \cline{2-3}
& Algorithm design& \citet{smith2014instance,vorraboot2012modified,campos2012local,brun2017framework}  \\ \cline{2-3}
& Algorithm understanding& \citet{zhao2016initial}  \\ \cline{2-3}
& Multiclass decomposition & \citet{LorenaCarvalho2010,quiterio2018using,moran2017use} \\
& & \citet{sun2019novel}\\
\cline{2-3}
& Parameter tuning& \citet{he2015quantification,nojima2011meta}  \\ \hline
\multirow{3}{*}{Meta-learning} & \multirow{3}{*}{Meta-features} & \citet{smith2014easy,leyvaset,GarciaEtAl2015b,garcia2018classifier,nojima2011meta} \\
& & \citet{sotoca2006meta,van2007measures,krijthe2012improving,cavalcanti2012data} \\ 
& & \citet{das2016meta,roy2016meta,parmezan2017metalearning,cruz2015meta,cruz2017meta} \\ 
& & \citet{munoz2018instance,ZHANG2019204,shah2018analyzing}\\
\hline 
\end{tabular}}
}
\end{table}

Concerning the usage of the individual measures, we can notice a variation per domain. As expected, feature-based measures are quite effective in FS. Among them, F1 is the most used and has shown highlighted results also in instance selection and in class imbalance analysis. LSC was proposed in the IS context. Neighborhood-based measures (mainly N1, N2 and N3) also show detached results in different domains, such as FS, noise identification and meta-learning. But one should be aware that in most of the reviewed work there was no clear evaluation on the contribution of each of the complexity measures values in the results achieved. Indeed, most of the related work perform an ad-hoc selection of which complexity measures are to be used (for example,  one representative measure per category). Since each measure provides a distinct perspective on classification complexity, a combination of different measures is advised. Nonetheless, whether there is a subset of the complexity measures that can be considered core to stress the difficulty of  problems from different application domains is still an open-issue.

%% file: Conclusion.tex
\section{Conclusion}\label{cap:concl}

This paper reviewed the main data complexity measures from the literature. These indices allow to characterize the difficulty of a classification problem from the perspectives of data geometry and distribution within or across the classes. They were first proposed and analyzed in \citet{HoBasu2002} and have since been extensively used in the analysis and development of classification and pre-processing techniques. 

The original complexity measures and other measures found in related literature were briefly presented. Despite the presence of many methods for measuring the complexity of classification problems, they often share similar concepts. There has not been a study comparing them to reveal which ones can extract more distinct aspects regarding data complexity. Besides the characteristics of each individual measure highlighted alongside their definitions, we present next some general discussions about each category of complexity measures.

    In the case of the feature-based complexity measures, there is an  expectation that each feature has a certain contribution to the discrimination task, and that the axis representing the feature can be interpreted as it is.  This is more likely to be true for problems where the features are meaningful explanatory variables each contributing somewhat independently to the classification.  It is particularly less likely to be true in  classification problems where sensory signals are directly taken as input, such as pixel values in images, where a natural unit of discriminatory information tends to involve a larger group of features (such as a patch of colors displayed over multiple pixels).   For those cases, transformation of the raw feature values, such as by a directional vector projection, becomes essential.  The second issue is that as we examine the overlap of the feature value ranges, there is an expectation that the unseen values in an interval that spans the seen values contribute to the discrimination task in a similar way as the seen values, i.e., there is continuity in the class definition w.r.t. that feature.  This tends to be true for features in a continuous numerical scale, and is less likely for other cases.  For categorical features, the notion of value ranges degenerates into specific values and several measures in this family have difficulties.

     The measures in the linearity family focus on the perspective of linear separability, which has a long history of being used as a characterization of classification difficulty.  It was involved in the early debates of the limits of certain classifier's capabilities (e.g. the debate on the perceptron in \citet{MinskyPapert1969}). One issue of concern is that linear separability is often characteristic of sparse data sets -- consider the extreme case where only one training point is available from each class in an arbitrary classification problem, and in that case linear separability of the training data does not give much information about the nature of the underlying task.  Sparse datasets in high dimensional space are also likely to be linearly separable (see, for example, \citep{costa2009using}), which motivates techniques like SVMs that use a feature transformation to map the data to a high dimensional space where simple linear classifiers suffice.  The interactive effects of this type of measures with data size, data density, and dimensionality are illustrative of the challenges involved in data complexity discussions.  Therefore the complexity evaluations need to be anchored first on fixed datasets, and followed by discussions of changes in responses to the other influences.

    Measures in this neighborhood-based family characterize the datasets in ways different from those of the feature-based family and the linearity-based family.  They use a distance function to summarize the relationship between points.  This is best fitted for datasets where the features are on a comparable scale (e.g. per-pixel intensity values) such that a natural metric exists.  For datasets that involve features of heterogeneous types and scales, a scale-normalization step or a suitable weighting scheme is needed for a summarizing metric to be properly defined.  The usefulness of the measures depends critically on whether such a metric can be obtained. The Gower distance metric employed in ECol is a simple alternative for dealing with features of different types and scales, but more sophisticate distance functions could be used instead \citep{wilson1997improved}.   In addition, since these measures are influenced by within-class data distributions as well as by the data distributions near the class boundaries, the information they convey may include more than what is relevant to the discrimination task, which may cause drown-out of the critical signal about classification complexity.

    The network based measures regard on the structure of the data in the input space. They may complement the previous measures presented, although they also consider the neighborhood of examples for obtaining the graph representation. It should be noticed that a number of other complex network measures can be extracted from the graph built, as well as other strategies can be used to obtain the graph representation. The strategy chosen to built the graph from a learning dataset considers both the proximity of the examples ($\epsilon$-NN) and the data label information (pruning step). Herewith, we expect to get an overview of both intra- and inter-class relationships.   
    
    All measures from the dimensionality group rely only on the numbers of examples and features in a dataset, disregarding the label information. Therefore, they do not give any indicative of boundary complexity, but rather give a very simplified and na\"ive overview on data sparsity. As discussed in the paper introduction, data sparsity is one of the factors that may affect the complexity of a classification problem.  Indeed, datasets with a high dimensionality and a low number of examples tend to be distributed sparsely. In many cases this can make the classification problem look simpler than it really is so that simple classification models may not generalize well to new data points that occupy regions formerly underrepresented in the training dataset. 
    
    The measures of the class imbalance category regard on the number of examples per class. As in the case of the dimensionality measures, they do not allow to directly estimate the complexity of the classification boundary. Rather, they regard on another aspect which may influence the performance of many ML classification techniques, which is the underrepresentation of one or more classes in relation to others.

This work also provides an R package with an implementation of a set of 22 complexity measures from the previous categories. The package is expected to give interested researchers a quick start into the field. An immediate line of follow-up work is to evaluate these measures empirically and try to: 
(i) identify those measures with most distinct concepts, since many of them have similar computation; 
and (ii) compare their ability in revealing the complexity of a diverse set of classification problems. This type of investigation is expected to yield a reduced subset of core measures able to capture the most critical aspects of classification complexity.

Lastly, the main use cases where the measures have been applied were presented. The most common use of the measures is to characterize datasets in meta-learning studies or the domain of competence of learning and pre-processing techniques. Nonetheless, more contributions remain possible in employing the conclusions of these studies to adapt and propose new learning and pre-processing techniques.  
For instance, relatively few works have been done in devising new learning schemes and pre-processing techniques based on the complexity measures. This points to the potentials of these measures that remain poorly explored.  We believe that a better understanding of the characteristics of a given problem shall be the key to support the design of techniques with better predictive results.

Another direction that awaits to be better explored is how one can use the complexity measures to evaluate different formulations of a problem, in terms of how classes are defined or chosen, in domains where there is flexibility in such choices. An example is a text categorization task, where one may have some limited freedom to choose what is to be considered a category.  Better choices can lead to lower error rates even if the classifier technology stays the same.  Here the data complexity measures can serve as figures of merit to evaluate alternative class definitions.

\section*{Acknowledgements}

This study was financed in part by the Coordena\c{c}\~{a}o de Aperfei\c{c}oamento de Pessoal de N\'{i}vel Superior - Brasil (CAPES) - Finance Code 001.
The authors would also like to thank the financial support of the foundations FAPESP (grant 2012/22608-8), CNPq (grants 308858/2014-0 and 305291/2017-3), CAPES (grant 88887.162551/2018-00) and CAPES-COFECUB.

%% file: main.bbl
\begin{thebibliography}{113}
\providecommand{\natexlab}[1]{#1}
\providecommand{\url}[1]{\texttt{#1}}
\expandafter\ifx\csname urlstyle\endcsname\relax
  \providecommand{\doi}[1]{doi: #1}\else
  \providecommand{\doi}{doi: \begingroup \urlstyle{rm}\Url}\fi

\bibitem[Ali and Smith(2006)]{ali2006learning}
Shawkat Ali and Kate~A Smith.
\newblock On learning algorithm selection for classification.
\newblock \emph{Applied Soft Computing}, 6\penalty0 (2):\penalty0 119--138,
  2006.

\bibitem[Anwar et~al.(2014)Anwar, Jones, and Ganesh]{anwar2014measurement}
Nafees Anwar, Geoff Jones, and Siva Ganesh.
\newblock Measurement of data complexity for classification problems with
  unbalanced data.
\newblock \emph{Statistical Analysis and Data Mining}, 7\penalty0 (3):\penalty0
  194--211, 2014.

\bibitem[Armano(2015)]{armano2015direct}
Giuliano Armano.
\newblock A direct measure of discriminant and characteristic capability for
  classifier building and assessment.
\newblock \emph{Information Sciences}, 325:\penalty0 466--483, 2015.

\bibitem[Armano and Tamponi(2016)]{armano2016experimenting}
Giuliano Armano and Emanuele Tamponi.
\newblock Experimenting multiresolution analysis for identifying regions of
  different classification complexity.
\newblock \emph{Pattern Analysis and Applications}, 19\penalty0 (1):\penalty0
  129--137, 2016.

\bibitem[Basu and Ho(2006)]{BasuHo2006}
Mitra Basu and Tin~K Ho.
\newblock \emph{Data complexity in pattern recognition}.
\newblock Springer, 2006.

\bibitem[Batista et~al.(2004)Batista, Prati, and Monard]{batista2004study}
Gustavo E A P~A Batista, Ronaldo~C Prati, and Maria~C Monard.
\newblock A study of the behavior of several methods for balancing machine
  learning training data.
\newblock \emph{ACM SIGKDD Explorations Newsletter}, 6\penalty0 (1):\penalty0
  20--29, 2004.

\bibitem[Baumgartner et~al.(2006)Baumgartner, Ho, Somorjai, Himmelreich, and
  Sorrell]{baumgartner2006complexity}
Richard Baumgartner, Tin~K Ho, Ray Somorjai, Uwe Himmelreich, and Tania
  Sorrell.
\newblock Complexity of magnetic resonance spectrum classification.
\newblock In \emph{Data Complexity in Pattern Recognition}, pages 241--248,
  2006.

\bibitem[Bernad{\'o}-Mansilla and Ho(2005)]{bernado2005domain}
Ester Bernad{\'o}-Mansilla and Tin~K Ho.
\newblock Domain of competence of xcs classifier system in complexity
  measurement space.
\newblock \emph{IEEE Transactions on Evolutionary Computation}, 9\penalty0
  (1):\penalty0 82--104, 2005.

\bibitem[Bottou and Lin(2007)]{bottou2007support}
L{\'e}on Bottou and Chih-Jen Lin.
\newblock Support vector machine solvers.
\newblock \emph{Large scale kernel machines}, 3\penalty0 (1):\penalty0
  301--320, 2007.

\bibitem[{Britto Jr} et~al.(2014){Britto Jr}, Sabourin, and
  Oliveira]{britto2014dynamic}
Alceu~S {Britto Jr}, Robert Sabourin, and Luiz E~S Oliveira.
\newblock Dynamic selection of classifiers - a comprehensive review.
\newblock \emph{Pattern Recognition}, 47\penalty0 (11):\penalty0 3665--3680,
  2014.

\bibitem[Brun et~al.(2018)Brun, {Britto Jr}, Oliveira, Enembreck, and
  Sabourin]{brun2017framework}
Andr{\'e}~L Brun, Alceu~S {Britto Jr}, Luiz~S Oliveira, Fabricio Enembreck, and
  Robert Sabourin.
\newblock A framework for dynamic classifier selection oriented by the
  classification problem difficulty.
\newblock \emph{Pattern Recognition}, 76:\penalty0 175--190, 2018.

\bibitem[Cali{\'n}ski and Harabasz(1974)]{calinski1974dendrite}
Tadeusz Cali{\'n}ski and Jerzy Harabasz.
\newblock A dendrite method for cluster analysis.
\newblock \emph{Communications in Statistics-theory and Methods}, 3\penalty0
  (1):\penalty0 1--27, 1974.

\bibitem[Campos et~al.(2012)Campos, Morell, and Ferri]{campos2012local}
Yoisel Campos, Carlos Morell, and Francesc~J Ferri.
\newblock A local complexity based combination method for decision forests
  trained with high-dimensional data.
\newblock In \emph{12th International Conference on Intelligent Systems Design
  and Applications (ISDA)}, pages 194--199, 2012.

\bibitem[Charte et~al.(2016)Charte, Rivera, {del Jesus}, and
  Herrera]{charte2016impact}
Francisco Charte, Antonio Rivera, Mar{\'\i}a~J {del Jesus}, and Francisco
  Herrera.
\newblock On the impact of dataset complexity and sampling strategy in
  multilabel classifiers performance.
\newblock In \emph{11th International Conference on Hybrid Artificial
  Intelligence Systems (HAIS)}, pages 500--511, 2016.

\bibitem[Ciarelli et~al.(2013)Ciarelli, Oliveira, and
  Salles]{ciarelli2013impact}
Patrick~M Ciarelli, Elias Oliveira, and Evandro O~T Salles.
\newblock Impact of the characteristics of data sets on incremental learning.
\newblock \emph{Artificial Intelligence Research}, 2\penalty0 (4):\penalty0
  63--74, 2013.

\bibitem[Costa et~al.(2009)Costa, Lorena, {y Peres}, and {de
  Souto}]{costa2009using}
Ivan~G Costa, Ana~C Lorena, Liciana R M~P {y Peres}, and Marcilio C~P {de
  Souto}.
\newblock Using supervised complexity measures in the analysis of cancer gene
  expression data sets.
\newblock In \emph{Brazilian Symposium on Bioinformatics}, pages 48--59, 2009.

\bibitem[Cristianini and Shawe-Taylor(2000)]{cristianini2000introduction}
Nello Cristianini and John Shawe-Taylor.
\newblock \emph{An introduction to support vector machines and other
  kernel-based learning methods}.
\newblock Cambridge university press, 2000.

\bibitem[Cruz et~al.(2015)Cruz, Sabourin, Cavalcanti, and Ren]{cruz2015meta}
Rafael M~O Cruz, Robert Sabourin, George D~C Cavalcanti, and Tsang~Ing Ren.
\newblock {META-DES}: A dynamic ensemble selection framework using
  meta-learning.
\newblock \emph{Pattern recognition}, 48\penalty0 (5):\penalty0 1925--1935,
  2015.

\bibitem[Cruz et~al.(2017{\natexlab{a}})Cruz, Sabourin, and
  Cavalcanti]{cruz2017meta}
Rafael M~O Cruz, Robert Sabourin, and George D~C Cavalcanti.
\newblock {META-DES.Oracle:} meta-learning and feature selection for dynamic
  ensemble selection.
\newblock \emph{Information fusion}, 38:\penalty0 84--103, 2017{\natexlab{a}}.

\bibitem[Cruz et~al.(2017{\natexlab{b}})Cruz, Zakane, Sabourin, and
  Cavalcanti]{cruz2017dynamic}
Rafael M~O Cruz, Hiba~H Zakane, Robert Sabourin, and George D~C Cavalcanti.
\newblock Dynamic ensemble selection vs k-nn: why and when dynamic selection
  obtains higher classification performance?
\newblock In \emph{17th International Conference on Image Processing Theory,
  Tools and Applications (IPTA)}, pages 1--6, 2017{\natexlab{b}}.

\bibitem[Cruz et~al.(2018)Cruz, Sabourin, and Cavalcanti]{cruz2018dynamic}
Rafael M~O Cruz, Robert Sabourin, and George D~C Cavalcanti.
\newblock Dynamic classifier selection: Recent advances and perspectives.
\newblock \emph{Information Fusion}, 41:\penalty0 195--216, 2018.

\bibitem[Cummins(2013)]{Cummins2013}
Lisa Cummins.
\newblock \emph{Combining and Choosing Case Base Maintenance Algorithms}.
\newblock PhD thesis, National University of Ireland, Cork, 2013.

\bibitem[Cummins and Bridge(2011)]{cummins2011dataset}
Lisa Cummins and Derek Bridge.
\newblock On dataset complexity for case base maintenance.
\newblock In \emph{19th International Conference on Case-Based Reasoning
  (ICCBR)}, pages 47--61. 2011.

\bibitem[{das D{\^o}res} et~al.(2016){das D{\^o}res}, Alves, Ruiz, and
  Barros]{das2016meta}
Silvia~N {das D{\^o}res}, Luciano Alves, Duncan~D Ruiz, and Rodrigo~C Barros.
\newblock A meta-learning framework for algorithm recommendation in software
  fault prediction.
\newblock In \emph{31st Annual ACM Symposium on Applied Computing (SAC)}, pages
  1486--1491, 2016.

\bibitem[{de Melo} and Lorena(2018)]{de2018using}
Vin{\'\i}cius~V {de Melo} and Ana~C Lorena.
\newblock Using complexity measures to evolve synthetic classification
  datasets.
\newblock In \emph{International Joint Conference on Neural Networks (IJCNN)},
  pages 1--8, 2018.

\bibitem[Dong and Kothari(2003)]{DongKothari2003}
Ming Dong and Rishabh~P Kothari.
\newblock Feature subset selection using a new definition of classificability.
\newblock \emph{Pattern Recognition Letters}, 24:\penalty0 1215--1225, 2003.

\bibitem[Elizondo et~al.(2012)Elizondo, Birkenhead, Gamez, Garcia, and
  Alfaro]{elizondo2012linear}
David~A Elizondo, Ralph Birkenhead, Matias Gamez, Noelia Garcia, and Esteban
  Alfaro.
\newblock Linear separability and classification complexity.
\newblock \emph{Expert Systems with Applications}, 39\penalty0 (9):\penalty0
  7796--7807, 2012.

\bibitem[Fern{\'a}ndez et~al.(2018)Fern{\'a}ndez, Garc{\'\i}a, Galar, Prati,
  Krawczyk, and Herrera]{fernandez2018learning}
Alberto Fern{\'a}ndez, Salvador Garc{\'\i}a, Mikel Galar, Ronaldo~C Prati,
  Bartosz Krawczyk, and Francisco Herrera.
\newblock \emph{Learning from imbalanced data sets}.
\newblock Springer, 2018.

\bibitem[Flores et~al.(2014)Flores, G{\'a}mez, and
  Mart{\'\i}nez]{flores2014domains}
Mar{\'i}a~J Flores, Jos{\'e}~A G{\'a}mez, and Ana~M Mart{\'\i}nez.
\newblock Domains of competence of the semi-naive bayesian network classifiers.
\newblock \emph{Information Sciences}, 260:\penalty0 120--148, 2014.

\bibitem[Fornells et~al.(2007)Fornells, Golobardes, Martorell, Garrell,
  Maci{\`a}, and Bernad{\'o}]{fornells2007methodology}
Albert Fornells, Elisabet Golobardes, Josep~M Martorell, Josep~M Garrell,
  N{\'u}ria Maci{\`a}, and Ester Bernad{\'o}.
\newblock A methodology for analyzing case retrieval from a clustered case
  memory.
\newblock In \emph{7th International Conference on Case-Based Reasoning
  (ICCBR)}, pages 122--136. 2007.

\bibitem[Frenay and Verleysen(2014)]{Frenay2013}
Benoit Frenay and Michel Verleysen.
\newblock Classification in the presence of label noise: a survey.
\newblock \emph{IEEE Transactions on Neural Networks and Learning Systems},
  25\penalty0 (5):\penalty0 845 -- 869, 2014.

\bibitem[Garcia et~al.(2013)Garcia, de~Carvalho, and Lorena]{garcia2013noisy}
Lu{\'\i}s P~F Garcia, Andr{\'e} C P L~F de~Carvalho, and Ana~C Lorena.
\newblock Noisy data set identification.
\newblock In \emph{8th International Conference on Hybrid Artificial
  Intelligent Systems (HAIS)}, pages 629--638. 2013.

\bibitem[Garcia et~al.(2015)Garcia, de~Carvalho, and Lorena]{GarciaEtAl2015}
Lu{\'\i}s P~F Garcia, Andr{\'e} C P L~F de~Carvalho, and Ana~C Lorena.
\newblock Effect of label noise in the complexity of classification problems.
\newblock \emph{Neurocomputing}, 160:\penalty0 108--119, 2015.

\bibitem[Garcia et~al.(2016)Garcia, de~Carvalho, and Lorena]{GarciaEtAl2015b}
Lu{\'\i}s P~F Garcia, Andr{\'e} C P L~F de~Carvalho, and Ana~C Lorena.
\newblock Noise detection in the meta-learning level.
\newblock \emph{Neurocomputing}, 176:\penalty0 14--25, 2016.

\bibitem[Garcia et~al.(2018)Garcia, Lorena, {de Souto}, and
  Ho]{garcia2018classifier}
Lu{\'\i}s P~F Garcia, Ana~C Lorena, Marcilio C~P {de Souto}, and Tin~Kam Ho.
\newblock Classifier recommendation using data complexity measures.
\newblock In \emph{24th International Conference on Pattern Recognition
  (ICPR)}, pages 874--879, 2018.

\bibitem[Garc{\'\i}a-Callejas and Ara{\'u}jo(2016)]{garcia2016effects}
David Garc{\'\i}a-Callejas and Miguel~B Ara{\'u}jo.
\newblock The effects of model and data complexity on predictions from species
  distributions models.
\newblock \emph{Ecological Modelling}, 326:\penalty0 4--12, 2016.

\bibitem[Garcia-Piquer et~al.(2012)Garcia-Piquer, Fornells, Orriols-Puig,
  Corral, and Golobardes]{garcia2012data}
Alvaro Garcia-Piquer, Albert Fornells, Albert Orriols-Puig, Guiomar Corral, and
  Elisabet Golobardes.
\newblock Data classification through an evolutionary approach based on
  multiple criteria.
\newblock \emph{Knowledge and information Systems}, 33\penalty0 (1):\penalty0
  35--56, 2012.

\bibitem[Gong and Huang(2012)]{gong2012kolmogorov}
Rongsheng Gong and Samuel~H Huang.
\newblock A kolmogorov--smirnov statistic based segmentation approach to
  learning from imbalanced datasets: With application in property refinance
  prediction.
\newblock \emph{Expert Systems with Applications}, 39\penalty0 (6):\penalty0
  6192--6200, 2012.

\bibitem[Gower(1971)]{gower1971general}
John Gower.
\newblock A general coefficient of similarity and some of its properties.
\newblock \emph{Biometrics}, 27\penalty0 (4):\penalty0 857--871, 1971.

\bibitem[He and Garcia(2009)]{he2009learning}
Haibo He and Edwardo~A Garcia.
\newblock Learning from imbalanced data.
\newblock \emph{IEEE Transactions on Knowledge and Data Engineering},
  21\penalty0 (9):\penalty0 1263--1284, 2009.

\bibitem[He et~al.(2015)He, Chan, Yeung, Pedrycz, and Ng]{he2015quantification}
Zhi-Min He, Patrick P~K Chan, Daniel~S Yeung, Witold Pedrycz, and Wing W~Y Ng.
\newblock Quantification of side-channel information leaks based on data
  complexity measures for web browsing.
\newblock \emph{International Journal of Machine Learning and Cybernetics},
  6\penalty0 (4):\penalty0 607--619, 2015.

\bibitem[Ho(2000)]{ho2000complexity}
Tin~K Ho.
\newblock Complexity of classification problems and comparative advantages of
  combined classifiers.
\newblock In \emph{Multiple Classifier Systems (MCS)}, pages 97--106. 2000.

\bibitem[Ho(2002)]{Ho2002}
Tin~K Ho.
\newblock A data complexity analysis of comparative advantages of decision
  forest constructors.
\newblock \emph{Pattern Analysis and Applications}, \penalty0 (5):\penalty0
  102--112, 2002.

\bibitem[Ho and Basu(2002)]{HoBasu2002}
Tin~K Ho and Mitra Basu.
\newblock Complexity measures of supervised classification problems.
\newblock \emph{IEEE Transactions on Pattern Analysis and Machine
  Intelligence}, 24\penalty0 (3):\penalty0 289--300, 2002.

\bibitem[Ho and Bernad{\'o}-Mansilla(2006)]{HoMansilla2006}
Tin~K Ho and Ester Bernad{\'o}-Mansilla.
\newblock Classifier domains of competence in data complexity space.
\newblock In \emph{Data complexity in pattern recognition}, pages 135--152,
  2006.

\bibitem[Ho et~al.(2006)Ho, Basu, and Law]{HoBasuLaw2006}
Tin~K Ho, Mitra Basu, and Martin H~C Law.
\newblock Measures of geometrical complexity in classification problems.
\newblock In \emph{Data Complexity in Pattern Recognition}, pages 1--23, 2006.

\bibitem[Hoekstra and Duin(1996)]{hoekstra1996nonlinearity}
Aarnoud Hoekstra and Robert P~W Duin.
\newblock On the nonlinearity of pattern classifiers.
\newblock In \emph{13th International Conference on Pattern Recognition
  (ICPR)}, volume~4, pages 271--275, 1996.

\bibitem[Hu et~al.(2010)Hu, Pedrycz, Yu, and Lang]{hu2010selecting}
Qinghua Hu, Witold Pedrycz, Daren Yu, and Jun Lang.
\newblock Selecting discrete and continuous features based on neighborhood
  decision error minimization.
\newblock \emph{IEEE Transactions on Systems, Man, and Cybernetics, Part B
  (Cybernetics)}, 40\penalty0 (1):\penalty0 137--150, 2010.

\bibitem[Kamath et~al.(2008)Kamath, Yeatman, and Eschrich]{kamath2008toward}
Vidya Kamath, Timothy~J Yeatman, and Steven~A Eschrich.
\newblock Toward a measure of classification complexity in gene expression
  signatures.
\newblock In \emph{30th Annual International Conference of the IEEE Engineering
  in Medicine and Biology Society (EMBS)}, pages 5704--5707, 2008.

\bibitem[Kim and Oommen(2009)]{kim2009using}
Sang-Woon Kim and John Oommen.
\newblock On using prototype reduction schemes to enhance the computation of
  volume-based inter-class overlap measures.
\newblock \emph{Pattern Recognition}, 42\penalty0 (11):\penalty0 2695--2704,
  2009.

\bibitem[Kolaczyk(2009)]{Kolaczyk2009}
Eric~D Kolaczyk.
\newblock \emph{Statistical Analysis of Network Data: Methods and Models}.
\newblock Springer Series in Statistics. Springer, 2009.

\bibitem[Kotsiantis and Kanellopoulos(2006)]{kotsiantis2006discretization}
Sotiris Kotsiantis and Dimitris Kanellopoulos.
\newblock Discretization techniques: A recent survey.
\newblock \emph{GESTS International Transactions on Computer Science and
  Engineering}, 32\penalty0 (1):\penalty0 47--58, 2006.

\bibitem[Krijthe et~al.(2012)Krijthe, Ho, and Loog]{krijthe2012improving}
Jesse~H Krijthe, Tin~K Ho, and Marco Loog.
\newblock Improving cross-validation based classifier selection using
  meta-learning.
\newblock In \emph{21st International Conference on Pattern Recognition
  (ICPR)}, pages 2873--2876, 2012.

\bibitem[Lebourgeois and Emptoz(1996)]{lebourgeois1996}
Frank Lebourgeois and Hubert Emptoz.
\newblock Pretopological approach for supervised learning.
\newblock In \emph{13th International Conference on Pattern Recognition},
  volume~4, pages 256--260, 1996.

\bibitem[Leyva et~al.(2014)Leyva, Gonz{\'a}lez, and P{\'e}rez]{leyvaset}
Enrique Leyva, Antonio Gonz{\'a}lez, and Ra{\'u}l P{\'e}rez.
\newblock A set of complexity measures designed for applying meta-learning to
  instance selection.
\newblock \emph{IEEE Transactions on Knowledge and Data Engineering},
  27\penalty0 (2):\penalty0 354--367, 2014.

\bibitem[Ling and Abu-Mostafa(2006)]{LiAbu-Mostafa2006}
Li~Ling and Yaser~S Abu-Mostafa.
\newblock Data complexity in machine learning.
\newblock Technical Report CaltechCSTR:2006.004, California Institute of
  Technology, 2006.

\bibitem[Liu et~al.(2010)Liu, Motoda, Setiono, and Zhao]{LiuEtAl2010}
Huan Liu, Hiroshi Motoda, Rudy Setiono, and Zheng Zhao.
\newblock Feature selection: An ever evolving frontier in data mining.
\newblock In \emph{4th International Workshop on Feature Selection in Data
  Mining (FSDM)}, volume~10, pages 4--13, 2010.

\bibitem[L{\'o}pez et~al.(2012)L{\'o}pez, Fern{\'a}ndez, Moreno-Torres, and
  Herrera]{lopez2012analysis}
Victoria L{\'o}pez, Alberto Fern{\'a}ndez, Jose~G Moreno-Torres, and Francisco
  Herrera.
\newblock Analysis of preprocessing vs. cost-sensitive learning for imbalanced
  classification. open problems on intrinsic data characteristics.
\newblock \emph{Expert Systems with Applications}, 39\penalty0 (7):\penalty0
  6585--6608, 2012.

\bibitem[Lorena and de~Carvalho(2010)]{LorenaCarvalho2010}
Ana~C Lorena and Andr{\'e} C P L~F de~Carvalho.
\newblock Building binary-tree-based multiclass classifiers using separability
  measures.
\newblock \emph{Neurocomputing}, 73\penalty0 (16-18):\penalty0 2837--2845,
  2010.

\bibitem[Lorena et~al.(2008)Lorena, de~Carvalho, and Gama]{LorenaEtAl2008}
Ana~C Lorena, Andr{\'e} C P L~F de~Carvalho, and Jo{\~a}o M~P Gama.
\newblock A review on the combination of binary classifiers in multiclass
  problems.
\newblock \emph{Artificial Intelligence Review}, 30\penalty0 (1):\penalty0
  19--37, 2008.

\bibitem[Lorena et~al.(2012)Lorena, Costa, Spola{\^o}r, and
  Souto]{LorenaEtAl2012}
Ana~C Lorena, Ivan~G Costa, Newton Spola{\^o}r, and Marcilio C~P Souto.
\newblock Analysis of complexity indices for classification problems: Cancer
  gene expression data.
\newblock \emph{Neurocomputing}, 75\penalty0 (1):\penalty0 33--42, 2012.

\bibitem[Lorena et~al.(2018)Lorena, Maciel, Miranda, Costa, and
  Prud{\^e}ncio]{lorena2018meta}
Ana~C Lorena, Aron~I Maciel, Pericles B~C Miranda, Ivan~G Costa, and Ricardo
  B~C Prud{\^e}ncio.
\newblock Data complexity meta-features for regression problems.
\newblock \emph{Machine Learning (accepted)}, 2018.

\bibitem[Lucca et~al.(2017)Lucca, Sanz, Dimuro, Bedregal, and
  Bustince]{lucca2017analyzing}
Giancarlo Lucca, Jose Sanz, Gra{\c{c}}aliz~P Dimuro, Benjam{\'\i}n Bedregal,
  and Humberto Bustince.
\newblock Analyzing the behavior of aggregation and pre-aggregation functions
  in fuzzy rule-based classification systems with data complexity measures.
\newblock In \emph{10th Conference of the European Society for Fuzzy Logic and
  Technology (IWIFSGN)}, pages 443--455, 2017.

\bibitem[Luengo and Herrera(2015)]{luengo2015automatic}
Juli{\'a}n Luengo and Francisco Herrera.
\newblock An automatic extraction method of the domains of competence for
  learning classifiers using data complexity measures.
\newblock \emph{Knowledge and Information Systems}, 42\penalty0 (1):\penalty0
  147--180, 2015.

\bibitem[Maci{\`a}(2011)]{Antolinez2011}
N{\'u}ria Maci{\`a}.
\newblock \emph{Data Complexity in Supervised Learning: a far-reaching
  implication}.
\newblock PhD thesis, La Salle, Universitat Ramon Llull, 2011.

\bibitem[Maci{\`a} and Bernad{\'o}-Mansilla(2014)]{macia2014towards}
N{\'u}ria Maci{\`a} and Ester Bernad{\'o}-Mansilla.
\newblock Towards uci+: A mindful repository design.
\newblock \emph{Information Sciences}, 261:\penalty0 237--262, 2014.

\bibitem[Macia et~al.(2008)Macia, Bernad{\'o}-Mansilla, and
  Orriols-Puig]{macia2008preliminary}
N{\'u}ria Macia, Ester Bernad{\'o}-Mansilla, and Albert Orriols-Puig.
\newblock Preliminary approach on synthetic data sets generation based on class
  separability measure.
\newblock In \emph{19th International Conference on Pattern Recognition
  (ICPR)}, pages 1--4, 2008.

\bibitem[Maci{\`a} et~al.(2010)Maci{\`a}, Orriols-Puig, and
  Bernad{\'o}-Mansilla]{macia2010search}
N{\'u}ria Maci{\`a}, Albert Orriols-Puig, and Ester Bernad{\'o}-Mansilla.
\newblock In search of targeted-complexity problems.
\newblock In \emph{12th annual conference on Genetic and evolutionary
  computation}, pages 1055--1062, 2010.

\bibitem[Maci{\`a} et~al.(2013)Maci{\`a}, Bernad{\'o}-Mansilla, Orriols-Puig,
  and Ho]{macia2013learner}
N{\'u}ria Maci{\`a}, Ester Bernad{\'o}-Mansilla, Albert Orriols-Puig, and
  Tin~Kam Ho.
\newblock Learner excellence biased by data set selection: A case for data
  characterisation and artificial data sets.
\newblock \emph{Pattern Recognition}, 46\penalty0 (3):\penalty0 1054--1066,
  2013.

\bibitem[Malina(2001)]{malina2001two}
Witold Malina.
\newblock Two-parameter fisher criterion.
\newblock \emph{IEEE Transactions on Systems, Man, and Cybernetics, Part B
  (Cybernetics)}, 31\penalty0 (4):\penalty0 629--636, 2001.

\bibitem[Ming and Vitanyi(1993)]{Vitanyi1993}
Li~Ming and Paul Vitanyi.
\newblock \emph{An Introduction to Kolmogorov Complexity and Its Applications}.
\newblock Springer, 1993.

\bibitem[Minsky and Papert(1969)]{MinskyPapert1969}
Marvin Minsky and Seymour Papert.
\newblock \emph{Perceptrons: An Introduction to Computational Geometry}.
\newblock The MIT Press, Cambridge MA, 1969.

\bibitem[Mollineda et~al.(2005)Mollineda, S{\'a}nchez, and
  Sotoca]{MollinedaEtAl2005}
Ram{\'o}n~A Mollineda, Jos{\'e}~S S{\'a}nchez, and Jos{\'e}~M Sotoca.
\newblock Data characterization for effective prototype selection.
\newblock In \emph{2nd Iberian Conference on Pattern Recognition and Image
  Analysis (IbPRIA)}, pages 27--34, 2005.

\bibitem[Mollineda et~al.(2006)Mollineda, S{\'a}nchez, and
  Sotoca]{sotoca2006meta}
Ram{\'o}n~A Mollineda, Jos{\'e}~S S{\'a}nchez, and Jos{\'e}~M Sotoca.
\newblock A meta-learning framework for pattern classification by means of data
  complexity measures.
\newblock \emph{Inteligencia Artificial}, 10\penalty0 (29):\penalty0 31--38,
  2006.

\bibitem[Morais and Prati(2013)]{Morais2013}
Gleison Morais and Ronaldo~C Prati.
\newblock Complex network measures for data set characterization.
\newblock In \emph{2nd Brazilian Conference on Intelligent Systems (BRACIS)},
  pages 12--18, 2013.

\bibitem[Mor{\'a}n-Fern{\'a}ndez
  et~al.(2017{\natexlab{a}})Mor{\'a}n-Fern{\'a}ndez, Bol{\'o}n-Canedo, and
  Alonso-Betanzos]{moran2017can}
Laura Mor{\'a}n-Fern{\'a}ndez, Ver{\'o}nica Bol{\'o}n-Canedo, and Amparo
  Alonso-Betanzos.
\newblock Can classification performance be predicted by complexity measures? a
  study using microarray data.
\newblock \emph{Knowledge and Information Systems}, 51\penalty0 (3):\penalty0
  1067--1090, 2017{\natexlab{a}}.

\bibitem[Mor{\'a}n-Fern{\'a}ndez
  et~al.(2017{\natexlab{b}})Mor{\'a}n-Fern{\'a}ndez, Bol{\'o}n-Canedo, and
  Alonso-Betanzos]{moran2017use}
Laura Mor{\'a}n-Fern{\'a}ndez, Ver{\'o}nica Bol{\'o}n-Canedo, and Amparo
  Alonso-Betanzos.
\newblock On the use of different base classifiers in multiclass problems.
\newblock \emph{Progress in Artificial Intelligence}, 6\penalty0 (4):\penalty0
  315--323, 2017{\natexlab{b}}.

\bibitem[Mthembu and Marwala(2008)]{mthembu2008note}
Linda Mthembu and Tshilidzi Marwala.
\newblock A note on the separability index.
\newblock \emph{arXiv preprint arXiv:0812.1107}, 2008.

\bibitem[Mu{\~n}oz et~al.(2018)Mu{\~n}oz, Villanova, Baatar, and
  Smith-Miles]{munoz2018instance}
Mario~A Mu{\~n}oz, Laura Villanova, Davaatseren Baatar, and Kate Smith-Miles.
\newblock Instance spaces for machine learning classification.
\newblock \emph{Machine Learning}, 107\penalty0 (1):\penalty0 109--147, 2018.

\bibitem[Nojima et~al.(2011)Nojima, Nishikawa, and Ishibuchi]{nojima2011meta}
Yusuke Nojima, Shinya Nishikawa, and Hisao Ishibuchi.
\newblock A meta-fuzzy classifier for specifying appropriate fuzzy partitions
  by genetic fuzzy rule selection with data complexity measures.
\newblock In \emph{IEEE International Conference on Fuzzy Systems (FUZZ)},
  pages 264--271, 2011.

\bibitem[Okimoto et~al.(2017)Okimoto, Savii, and Lorena]{okimoto2017}
Lucas~Chesini Okimoto, Ricardo~Manh{\~a}es Savii, and Ana~Carolina Lorena.
\newblock Complexity measures effectiveness in feature selection.
\newblock In \emph{6th Brazilian Conference on Intelligent Systems (BRACIS)},
  pages 91--96, 2017.

\bibitem[Orriols-Puig et~al.(2010)Orriols-Puig, Maci{\`a}, and
  Ho]{Orriols-PuigEtAl2010}
Albert Orriols-Puig, N{\'u}ria Maci{\`a}, and Tin~K Ho.
\newblock Documentation for the data complexity library in c++.
\newblock Technical report, La Salle - Universitat Ramon Llull, 2010.

\bibitem[Parmezan et~al.(2017)Parmezan, Lee, and Wu]{parmezan2017metalearning}
Antonio R~S Parmezan, Huei~D Lee, and Feng~C Wu.
\newblock Metalearning for choosing feature selection algorithms in data
  mining: Proposal of a new framework.
\newblock \emph{Expert Systems with Applications}, 75:\penalty0 1--24, 2017.

\bibitem[Pranckeviciene et~al.(2006)Pranckeviciene, Ho, and
  Somorjai]{PranckevicieneEtAl2006}
Erinija Pranckeviciene, Tin~K Ho, and Ray Somorjai.
\newblock Class separability in spaces reduced by feature selection.
\newblock In \emph{18th International Conference on Pattern Recognition
  (ICPR)}, volume~2, pages 254--257, 2006.

\bibitem[Quiterio and Lorena(2018)]{quiterio2018using}
Thaise~M Quiterio and Ana~C Lorena.
\newblock Using complexity measures to determine the structure of directed
  acyclic graphs in multiclass classification.
\newblock \emph{Applied Soft Computing}, 65:\penalty0 428--442, 2018.

\bibitem[Ren and Vale(2012)]{cavalcanti2012data}
George D C Cavalcantiand Tsang~I Ren and Breno~A Vale.
\newblock Data complexity measures and nearest neighbor classifiers: A
  practical analysis for meta-learning.
\newblock In \emph{24th International Conference on Tools with Artificial
  Intelligence (ICTAI)}, volume~1, pages 1065--1069, 2012.

\bibitem[Roy et~al.(2016)Roy, Cruz, Sabourin, and Cavalcanti]{roy2016meta}
Anandarup Roy, Rafael M~O Cruz, Robert Sabourin, and George D~C Cavalcanti.
\newblock Meta-learning recommendation of default size of classifier pool for
  {META-DES}.
\newblock \emph{Neurocomputing}, 216:\penalty0 351--362, 2016.

\bibitem[Sa{\'e}z et~al.(2013)Sa{\'e}z, Luengo, and Herrera]{SaezEtAl2013}
Jos{\'e}~A Sa{\'e}z, Juli{\'a}n Luengo, and Francisco Herrera.
\newblock Predicting noise filtering efficacy with data complexity measures for
  nearest neighbor classification.
\newblock \emph{Pattern Recognition}, 46\penalty0 (1):\penalty0 355--364, 2013.

\bibitem[Santos et~al.(2018)Santos, Soares, Abreu, Araujo, and
  Santos]{santos2018cross}
Miriam~Seoane Santos, Jastin~Pompeu Soares, Pedro~Henrigues Abreu, Helder
  Araujo, and Joao Santos.
\newblock Cross-validation for imbalanced datasets: Avoiding overoptimistic and
  overfitting approaches.
\newblock \emph{IEEE Computational Intelligence Magazine}, 13\penalty0
  (4):\penalty0 59--76, 2018.

\bibitem[Seijo-Pardo et~al.(2019)Seijo-Pardo, Bol{\'o}n-Canedo, and
  Alonso-Betanzos]{seijo2019developing}
Borja Seijo-Pardo, Ver{\'o}nica Bol{\'o}n-Canedo, and Amparo Alonso-Betanzos.
\newblock On developing an automatic threshold applied to feature selection
  ensembles.
\newblock \emph{Information Fusion}, 45:\penalty0 227--245, 2019.

\bibitem[Shah et~al.(2018)Shah, Khemani, Azarian, Pecht, and
  Su]{shah2018analyzing}
Rushit Shah, Varun Khemani, Michael Azarian, Michael Pecht, and Yan Su.
\newblock Analyzing data complexity using metafeatures for classification
  algorithm selection.
\newblock In \emph{Prognostics and System Health Management Conference
  (PHM-Chongqing)}, pages 1280--1284, 2018.

\bibitem[Singh(2003{\natexlab{a}})]{Singh2003}
Sameer Singh.
\newblock Multiresolution estimates of classification complexity.
\newblock \emph{IEEE Transactions on Pattern Analysis and Machine
  Intelligence}, 25\penalty0 (12):\penalty0 1534--1539, 2003{\natexlab{a}}.

\bibitem[Singh(2003{\natexlab{b}})]{Singh2003PRISM}
Sameer Singh.
\newblock {PRISM:} a novel framework for pattern recognition.
\newblock \emph{Pattern Analysis and Applications}, 6\penalty0 (2):\penalty0
  134--149, 2003{\natexlab{b}}.

\bibitem[Skrypnyk(2011)]{skrypnyk2011irrelevant}
Iryna Skrypnyk.
\newblock Irrelevant features, class separability, and complexity of
  classification problems.
\newblock In \emph{23rd IEEE International Conference on Tools with Artificial
  Intelligence (ICTAI)}, pages 998--1003, 2011.

\bibitem[Smith(1968)]{smith1968pattern}
Fred~W Smith.
\newblock Pattern classifier design by linear programming.
\newblock \emph{IEEE Transactions on Computers}, C-17\penalty0 (4):\penalty0
  367--372, 1968.

\bibitem[Smith et~al.(2014{\natexlab{a}})Smith, Martinez, and
  Giraud-Carrier]{smith2014instance}
Michael~R Smith, Tony Martinez, and Christophe Giraud-Carrier.
\newblock An instance level analysis of data complexity.
\newblock \emph{Machine Learning}, 95\penalty0 (2):\penalty0 225--256,
  2014{\natexlab{a}}.

\bibitem[Smith et~al.(2014{\natexlab{b}})Smith, White, Giraud-Carrier, and
  Martinez]{smith2014easy}
Michael~R Smith, Andrew White, Christophe Giraud-Carrier, and Tony Martinez.
\newblock An easy to use repository for comparing and improving machine
  learning algorithm usage.
\newblock \emph{arXiv preprint arXiv:1405.7292}, 2014{\natexlab{b}}.

\bibitem[Smith-Miles(2009)]{smith2009cross}
Kate~A Smith-Miles.
\newblock Cross-disciplinary perspectives on meta-learning for algorithm
  selection.
\newblock \emph{ACM Computing Surveys (CSUR)}, 41\penalty0 (1):\penalty0 1--26,
  2009.

\bibitem[Sotoca et~al.(2005)Sotoca, S{\'a}nchez, and Mollineda]{SotocaEtAl2005}
Jos{\'e}~M Sotoca, Jos{\'e} S{\'a}nchez, and Ram{\'o}n~A Mollineda.
\newblock A review of data complexity measures and their applicability to
  pattern classification problems.
\newblock In \emph{Actas del III Taller Nacional de Miner{\'i}a de Dados y
  Aprendizaje (TAMIDA)}, pages 77--83, 2005.

\bibitem[Souto et~al.(2010)Souto, Lorena, Spola{\^o}r, and
  Costa]{SoutoEtAl2010}
Marcilio C~P Souto, Ana~C Lorena, Newton Spola{\^o}r, and Ivan~G Costa.
\newblock Complexity measures of supervised classification tasks: a case study
  for cancer gene expression data.
\newblock In \emph{International Joint Conference on Neural Networks (IJCNN)},
  pages 1352--1358, 2010.

\bibitem[Sun et~al.(2019)Sun, Liu, Wu, Hong, Wang, and Zhang]{sun2019novel}
MengXin Sun, KunHong Liu, QingQiang Wu, QingQi Hong, BeiZhan Wang, and Haiying
  Zhang.
\newblock A novel {ECOC} algorithm for multiclass microarray data
  classification based on data complexity analysis.
\newblock \emph{Pattern Recognition}, 90:\penalty0 346--362, 2019.

\bibitem[Tanwani and Farooq(2010)]{tanwani2010classification}
Ajay~K Tanwani and Muddassar Farooq.
\newblock Classification potential vs. classification accuracy: a comprehensive
  study of evolutionary algorithms with biomedical datasets.
\newblock \emph{Learning Classifier Systems}, 6471:\penalty0 127--144, 2010.

\bibitem[Trujillo et~al.(2011)Trujillo, Mart{\'i}nez, Galv{\'a}n-L{\'o}pez, and
  Legrand]{TrujilloEtAl2011}
Leonardo Trujillo, Yuliana Mart{\'i}nez, Edgar Galv{\'a}n-L{\'o}pez, and
  Pierrick Legrand.
\newblock Predicting problem difficulty for genetic programming applied to data
  classification.
\newblock In \emph{13th annual conference on Genetic and evolutionary
  computation (GECCO)}, pages 1355--1362, 2011.

\bibitem[Vilalta and Drissi(2002)]{VilaltaDrissi2002}
Ricardo Vilalta and Youssef Drissi.
\newblock A perspective view and survey of meta-learning.
\newblock \emph{Artificial Intelligence Review}, 18\penalty0 (2):\penalty0
  77--95, 2002.

\bibitem[Vorraboot et~al.(2012)Vorraboot, Rasmequan, Lursinsap, and
  Chinnasarn]{vorraboot2012modified}
Piyanoot Vorraboot, Suwanna Rasmequan, Chidchanok Lursinsap, and Krisana
  Chinnasarn.
\newblock A modified error function for imbalanced dataset classification
  problem.
\newblock In \emph{7th International Conference on Computing and Convergence
  Technology (ICCCT)}, pages 854--859, 2012.

\bibitem[Walt and Barnard(2007)]{van2007measures}
Christiaan V~D Walt and Etienne Barnard.
\newblock Measures for the characterisation of pattern-recognition data sets.
\newblock In \emph{18th Annual Symposium of the Pattern Recognition Association
  of South Africa (PRASA)}, 2007.

\bibitem[Wilson and Martinez(1997)]{wilson1997improved}
D~Randall Wilson and Tony~R Martinez.
\newblock Improved heterogeneous distance functions.
\newblock \emph{Journal of artificial intelligence research}, 6:\penalty0
  1--34, 1997.

\bibitem[Wolpert(1996)]{wolpert1996lack}
David~H Wolpert.
\newblock The lack of a priori distinctions between learning algorithms.
\newblock \emph{Neural computation}, 8\penalty0 (7):\penalty0 1341--1390, 1996.

\bibitem[Xing et~al.(2013)Xing, Cai, Cai, Hejlesen, and
  Toft]{xing2013preliminary}
Yan Xing, Hao Cai, Yanguang Cai, Ole Hejlesen, and Egon Toft.
\newblock Preliminary evaluation of classification complexity measures on
  imbalanced data.
\newblock In \emph{Chinese Intelligent Automation Conference: Intelligent
  Information Processing"}, pages 189--196, 2013.

\bibitem[Zhang et~al.(2019)Zhang, Li, Zhang, Yang, Guo, and Ji]{ZHANG2019204}
Xueying Zhang, Ruixian Li, Bo~Zhang, Yunxiang Yang, Jing Guo, and Xiang Ji.
\newblock An instance-based learning recommendation algorithm of imbalance
  handling methods.
\newblock \emph{Applied Mathematics and Computation}, 351:\penalty0 204 -- 218,
  2019.

\bibitem[Zhao et~al.(2016)Zhao, Cao, Zhu, Ming, and Ashfaq]{zhao2016initial}
Xingmin Zhao, Weipeng Cao, Hongyu Zhu, Zhong Ming, and Rana Aamir~Raza Ashfaq.
\newblock An initial study on the rank of input matrix for extreme learning
  machine.
\newblock \emph{International Journal of Machine Learning and Cybernetics},
  pages 1--13, 2016.

\bibitem[Zhu et~al.(2005)Zhu, Lafferty, and Rosenfeld]{Zhu2005}
Xiaojin Zhu, John Lafferty, and Ronald Rosenfeld.
\newblock \emph{Semi-supervised learning with graphs}.
\newblock PhD thesis, Carnegie Mellon University, Language Technologies
  Institute, School of Computer Science, 2005.

\bibitem[Zubek and Plewczynski(2016)]{zubek2016complexity}
Julian Zubek and Dariusz~M Plewczynski.
\newblock Complexity curve: a graphical measure of data complexity and
  classifier performance.
\newblock \emph{PeerJ Computer Science}, 2:\penalty0 e76, 2016.

\end{thebibliography}
